\documentclass{article}
\usepackage{setspace,etoolbox}
\AtBeginEnvironment{table}{\singlespacing}
\usepackage{enumitem} 
\setlist[enumerate]{font=\sffamily} 

\usepackage{color,soul}

\usepackage{natbib}
\bibpunct[, ]{(}{)}{,}{a}{}{,}%

\usepackage{xcolor}

\usepackage{threeparttable}
\usepackage{tcolorbox}
\usepackage{adjustbox}
\usepackage{booktabs}
\usepackage{pgfplots}
\pgfplotsset{compat=1.18}
\usepgfplotslibrary{statistics, groupplots}
\pgfplotsset{
  colormap={divRedWhiteBlue}{
    color(0cm)=(red);
    color(0.5cm)=(white);
    color(1cm)=(blue)
  }
}
\pgfdeclarelayer{background}
\pgfsetlayers{background,main}
\usepackage{graphicx}
\usepackage{amsmath}
\usepackage{amssymb}
\usepackage{amsthm}
\usepackage{caption}
\usepackage{bigints}
\usepackage{lmodern}
\usepackage{tikz}
\usetikzlibrary{fit,trees,shapes,decorations,spy,calc,positioning,arrows.meta,backgrounds}
\usepackage[breaklinks]{hyperref}
\hypersetup{
colorlinks = {true}, citecolor = {blue}, menubordercolor = {red}, citebordercolor = {white}, linkbordercolor = {red}, urlcolor = {black} }

\newcommand{\tokenizertkn}[2]{\adjustbox{valign=M, margin=0pt}{\colorbox{#1}{\strut#2}}}

\newtheorem{proposition}{Proposition}
\newtheorem{lemma}{Lemma}
\newtheorem{corollary}{Corollary}
\newtheorem{assumption}{Assumption}
\theoremstyle{remark}

 \DeclareMathOperator*{\argmin}{arg\,min}

\newcommand{\E}{\mathbb{E}}

\newcommand{\one}{\mathbf{1}}
\newcommand{\Pa}{\Pi}
\newcommand{\vlow}{\underline{v}}
\newcommand{\vhigh}{\overline{v}}

\newcommand{\Vone}{V_{(1)}}
\newcommand{\Vtwo}{V_{(2)}}

\newcommand{\ResNDB}{32,853}  
\newcommand{\ResCStar}{0.398}  
\newcommand{\ResMetTargetOne}{78.8}  
\newcommand{\ResMedianReserve}{18,201}  
\newcommand{\ResReserveQlo}{11,480}  
\newcommand{\ResReserveQhi}{29,741}  
\newcommand{\ResMedianWidth}{675}  
\newcommand{\ResMedianWidthShare}{3.8}  
\newcommand{\ResSupportBelow}{99}  
\newcommand{\ResSupportAbove}{0}  
\newcommand{\ResXiMedian}{0.143}  
\newcommand{\ResXiPositiveShare}{99.96}  
\newcommand{\ResAnchorRsq}{0.634}  
\newcommand{\ResAnchorRsqPre}{0.296}  
\newcommand{\ResMarkupQlo}{1.06}  
\newcommand{\ResMarkupQhi}{1.29}  
\newcommand{\ResInterceptShareQlo}{0.05}  
\newcommand{\ResInterceptShareQhi}{0.18}  
\newcommand{\ResCorrInterceptAnchor}{$-$0.71}  
\newcommand{\ResCorrXiAnchor}{0.008}  
\newcommand{\ResMakeLowest}{Porsche}  
\newcommand{\ResMakeLowestXi}{0.116}  
\newcommand{\ResMakeLowestMarkup}{1.131}  
\newcommand{\ResMakeHighest}{Jaguar}  
\newcommand{\ResMakeHighestXi}{0.180}  
\newcommand{\ResMakeHighestMarkup}{1.220}  
\newcommand{\ResBetweenMakeShare}{8.1}  
\newcommand{\ResFeaturedReserve}{18,149}  
\newcommand{\ResPanelAN}{400}  
\newcommand{\ResScheduleMedianCzero}{3,808}  
\newcommand{\ResScheduleMetCzero}{99.8}  
\newcommand{\ResScheduleMedianCone}{39,920}  
\newcommand{\ResScheduleMetCone}{22.9}  
\newcommand{\ResBandNarrowInShare}{88.5}  
\newcommand{\ResBandNarrowBelow}{2,866}  
\newcommand{\ResBandNarrowAbove}{898}  
\newcommand{\ResBandNoneTwoEightInShare}{98.8}  
\newcommand{\ResBandNtwoZeroFourEightInShare}{99.9}  
\newcommand{\ResBandMedianLower}{6,228}  
\newcommand{\ResOwnThresholdMedian}{19,000}  
\newcommand{\ResReserveBelowOwnThresholdShare}{55.6}  
\newcommand{\ResHeldOutDiff}{0.0132}  
\newcommand{\ResHeldOutT}{0.4}  
\newcommand{\ResGapFT}{0.125}  
\newcommand{\ResGapPT}{0.020}  
\newcommand{\ResDollarMAEFT}{4,522}  
\newcommand{\ResDollarMAEPT}{5,080}  
\newcommand{\ResDollarOverFT}{17.0}  
\newcommand{\ResDollarOverPT}{20.0}  
\newcommand{\ResDollarPairedMean}{446}  
\newcommand{\ResDollarPairedSE}{31}  
\newcommand{\ResDollarShareCloser}{59.6}  
\newcommand{\ResDollarFTConverged}{33}  
\newcommand{\ResDollarFTStarts}{34}  
\newcommand{\ResDollarFTGateStatus}{fails the domain gate}  
\newcommand{\ResDollarFTHoldoutViolations}{1}  
\newcommand{\ResDollarFTHoldoutRows}{32,853}  
\newcommand{\ResDollarPTConverged}{34}  
\newcommand{\ResDollarPTStarts}{34}  
\newcommand{\ResDollarPTGateStatus}{passes every production gate}  
\newcommand{\ResWtwoSpearmanK}{0.104}  
\newcommand{\ResWtwoSpearmanCount}{0.041}  
\newcommand{\ResCountDepthCorr}{0.626}  
\newcommand{\ResGradientGap}{$2.2\times10^{-16}$}  
\newcommand{\ResAttrN}{512}  
\newcommand{\ResFeaturedTitle}{2001 Porsche 911 Turbo 6-Speed}  
\newcommand{\ResFeaturedTokens}{682}  
\newcommand{\ResFeaturedMaxPhi}{5,259}  
\newcommand{\ResFeaturedMaxPhiRaw}{5258.897273}  
\newcommand{\ResFeaturedHalfMaxPhi}{2,629}  
\newcommand{\ResFeaturedHalfMaxPhiRaw}{2629.448636}  
\newcommand{\ResFeaturedSumPhi}{127,827}  
\newcommand{\ResFeaturedTopPosPhi}{2,730}  
\newcommand{\ResFeaturedTopNegPhi}{$-$5,259}  
\newcommand{\ResAttrWordsQualifying}{971}  
\newcommand{\ResAttrWordsMinListings}{20}  
\newcommand{\ResAttrLowerTopWord}{Reserve}  
\newcommand{\ResAttrLowerTopPct}{18.3}  
\newcommand{\ResAttrLowerTopListings}{106}  
\newcommand{\ResAttrLowerSecondPct}{6.9}  
\newcommand{\ResAttrLowerLastPct}{3.4}  
\newcommand{\ResAttrRaiseTopPct}{7.6}  
\newcommand{\ResAttrRaiseLastPct}{4.5}  
\newcommand{\ResGradN}{11,066}  
\newcommand{\ResGradMinGroup}{20}  
\newcommand{\ResGradBtwo}{$-$0.178}  
\newcommand{\ResGradBtwoSE}{0.026}  
\newcommand{\ResGradDeclineBtwo}{27.4}  
\newcommand{\ResGradReserve}{$-$0.126}  
\newcommand{\ResGradDeclineReserve}{20.2}  
\newcommand{\ResGradPsi}{$-$0.056}  
\newcommand{\ResGradMarkup}{$-$0.007}  
\newcommand{\ResGradShare}{70}  
\newcommand{\ResGradStepHundredBtwo}{20.0}  
\newcommand{\ResGradStepHundredReserve}{10.9}  
\newcommand{\ResStageTwoObjective}{-2.5752}  
\newcommand{\ResStageTwoConverged}{35}  
\newcommand{\ResStageTwoStarts}{35}  
\newcommand{\ResStageTwoSpread}{$3.0\times10^{-14}$}  
\newcommand{\ResStageTwoReserveMovement}{$8.2\times10^{-7}$}  
\newcommand{\ResStageTwoAuditRows}{512}  
\newcommand{\ResJacobianCond}{$5.1\times10^{6}$}  
\newcommand{\ResScoreGap}{$1.9\times10^{-16}$}  
\newcommand{\ResNDA}{32,862}  
\newcommand{\ResStageOneBest}{-2.5669}  
\newcommand{\ResStageOneBestEpoch}{5}  
\newcommand{\ResStageOneHalfLRLoss}{0.007}  
\newcommand{\ResStageOneRankLoss}{0.007}  
\newcommand{\ResWtwoCountLow}{0.600}  
\newcommand{\ResWtwoCountHigh}{0.639}  
\newcommand{\ResBandNarrowLower}{11,042}  
\newcommand{\ResBandNarrowUpper}{28,788}  
\newcommand{\ResBandWideInShare}{99.7}  
\newcommand{\ResSupportVzeroShare}{94.9}  
\newcommand{\ResPolNLow}{2,843}  
\newcommand{\ResPolMedianGain}{868}  
\newcommand{\ResPolTotalGain}{6.0}  
\newcommand{\ResPolGainShare}{9.8}  
\newcommand{\ResPolNHigh}{4,796}  
\newcommand{\ResPolHighBidTotal}{285.1}  
\newcommand{\ResPolLowRevenue}{60.8}  
\newcommand{\ResPolNetRevenue}{224.3}  
\newcommand{\ResPolIncumbentRevenueBillion}{1.134}  
\newcommand{\ResPolNetIncumbentShare}{19.8}  
\newcommand{\ResPolTotalCLow}{3.4}  
\newcommand{\ResPolTotalCHigh}{10.1}  

\renewcommand{\ResCStar}{0.467} 
\renewcommand{\ResMetTargetOne}{72.3} 
\renewcommand{\ResMedianReserve}{20,686} 
\renewcommand{\ResReserveQlo}{12,978} 
\renewcommand{\ResReserveQhi}{33,947} 
\renewcommand{\ResMedianWidth}{694} 
\renewcommand{\ResMedianWidthShare}{3.5} 
\renewcommand{\ResSupportBelow}{38} 
\renewcommand{\ResSupportAbove}{3} 
\renewcommand{\ResScheduleMetCzero}{99.9} 
\renewcommand{\ResScheduleMetCone}{24.0} 
\renewcommand{\ResBandNarrowInShare}{87.9} 
\renewcommand{\ResBandNarrowBelow}{1,330} 
\renewcommand{\ResBandNarrowAbove}{2,644} 
\renewcommand{\ResBandNoneTwoEightInShare}{99.1} 
\renewcommand{\ResBandNtwoZeroFourEightInShare}{100.0} 
\renewcommand{\ResReserveBelowOwnThresholdShare}{46.6} 
\renewcommand{\ResHeldOutDiff}{0.0495} 
\renewcommand{\ResHeldOutT}{17.2} 
\renewcommand{\ResGapFT}{0.089} 
\renewcommand{\ResDollarFTHoldoutViolations}{2} 
\renewcommand{\ResGradReserve}{$-$0.128} 
\renewcommand{\ResGradDeclineReserve}{20.4} 
\renewcommand{\ResGradShare}{72} 
\renewcommand{\ResGradStepHundredReserve}{11.0} 
\renewcommand{\ResBandWideInShare}{99.9} 
\renewcommand{\ResSupportVzeroShare}{97.8} 
\renewcommand{\ResPolNLow}{4,194} 
\renewcommand{\ResPolMedianGain}{1,239} 
\renewcommand{\ResPolTotalGain}{11.9} 
\renewcommand{\ResPolGainShare}{11.8} 
\renewcommand{\ResPolNHigh}{4,194} 
\renewcommand{\ResPolHighBidTotal}{266.7} 
\renewcommand{\ResPolLowRevenue}{101.0} 
\renewcommand{\ResPolNetRevenue}{165.8} 
\renewcommand{\ResPolNetIncumbentShare}{14.6} 
\renewcommand{\ResPolTotalCLow}{6.1} 
\renewcommand{\ResPolTotalCHigh}{15.8} 
\newcommand{\ResPolicyEligible}{25,173}
\newcommand{\ResPolicyNoReserve}{7,680}

\newcommand{\ResComparisonValidN}{32,851}

\usepackage{arxiv}
\usepackage{microtype}
\newcommand{\PaperTitle}{Setting Optimal Reserve Prices using Fine-Tuned Language Models}
\newcommand{\PaperAbstract}{\textbf{Problem definition.} Online auction houses must set reserve prices for highly differentiated products whose unstructured descriptions contain valuable demand information. Existing approaches generally require managers to specify product attributes in advance. We ask whether a language model can instead learn from listing descriptions and historical bids to set item-specific reserve prices before bidding begins. \\
\textbf{Methodology/results.} We fine-tune a pretrained text embedding language model using the structural likelihood of auction bids as its training objective. This directs representation learning toward the valuation distribution that determines the reserve decision. We show that consecutive upper bids identify the tail information needed to set an optimal reserve without observing or modeling the number of potential bidders. A split-sample estimator learns the representation on one half of the data and conducts structural estimation and inference conditional on that representation on the other. We apply the method to approximately 65,800 online used-vehicle auctions. With seller values calibrated to match the sale rate among listings, the model implies a counterfactual net increase of \$\ResPolNetRevenue{} million, representing a \ResPolNetIncumbentShare\% increase above the incumbent reserve policy. \\
\textbf{Managerial implications.} The framework translates a listing description and a seller's value into an item-specific reserve recommendation without constructing a product taxonomy or requiring a bid history for the item being priced. It provides an approach to adapting language models for operational decisions in marketplaces where products are heterogeneous and demand-relevant information is recorded in text.}
\newcommand{\PaperKeywords}{Auctions; Reserve price; Structural estimation; Text as data; Language models; Revenue management}
\newcommand{\PaperVersionDate}{November 12, 2026}

\renewcommand{\shorttitle}{\PaperTitle}
\title{\PaperTitle}
\author{Joshua Foster and Fredrik {\O}degaard\\[0.75ex]
  \normalsize Ivey Business School, Western University}
\date{\PaperVersionDate}
\hypersetup{pdftitle={\PaperTitle},pdfauthor={Joshua Foster and Fredrik Odegaard},
  pdfkeywords={\PaperKeywords}}

\begin{document}
\maketitle
\begin{abstract}
\PaperAbstract
\end{abstract}
\newpage

\section{Introduction}\label{sec:introduction}

Auctions provide a fast and efficient selling mechanism that has been employed for millennia. Classic applications encompass a variety of markets such as those for tuna fish in Tokyo, fine art and antiques in London, and flowers in Holland. In this paper, we consider the modern counterpart of online auction platforms \citep{pinker2003managing,overby2009electronic} that serve as an important market mechanism for individual sellers seeking to maximize revenue for a wide variety of objects. 

A foundational problem in this pursuit concerns the reserve price $r$ of the seller. This reserve establishes a price threshold that at least one bid must exceed in order for the auction to be successfully concluded. Established independently and contemporaneously by
\cite{riley1981optimal} and \cite{myerson1981optimal}, the seller's optimal reserve price $r^*$ can be expressed as the solution to
\begin{equation}
  r^* = v_0 + \frac{1 - F(r^*)}{f(r^*)},
  \label{eq:myerson1}
\end{equation}
where $v_0$ denotes the seller's own valuation and $F(\cdot)$ and $f(\cdot)$ denote the distribution and density of a bidder's private valuation $V$. A defining feature of this conditional is that the optimal reserve is invariant to the number of bidders \citep{EngelbrechtWiggans1987}, since it directly equates the marginal revenue of the marginal type to the seller's opportunity cost.

While theoretically prescriptive, this condition is frequently difficult to implement because the latent valuation distribution $F$ is rarely observable in practice. Nonetheless, in stable market environments featuring frequent transactions of homogeneous goods (e.g., wholesale fish auctions), inferring the functional form of $F$ from historical transactional data can be relatively straightforward. Conversely, in markets characterized by significant product differentiation and unobserved heterogeneity, estimating this valuation distribution poses a substantial empirical challenge, typically necessitating \emph{ad hoc}, listing-specific structural assumptions.

\begin{figure}[htbp]
  \centering
  \includegraphics[width=\textwidth]{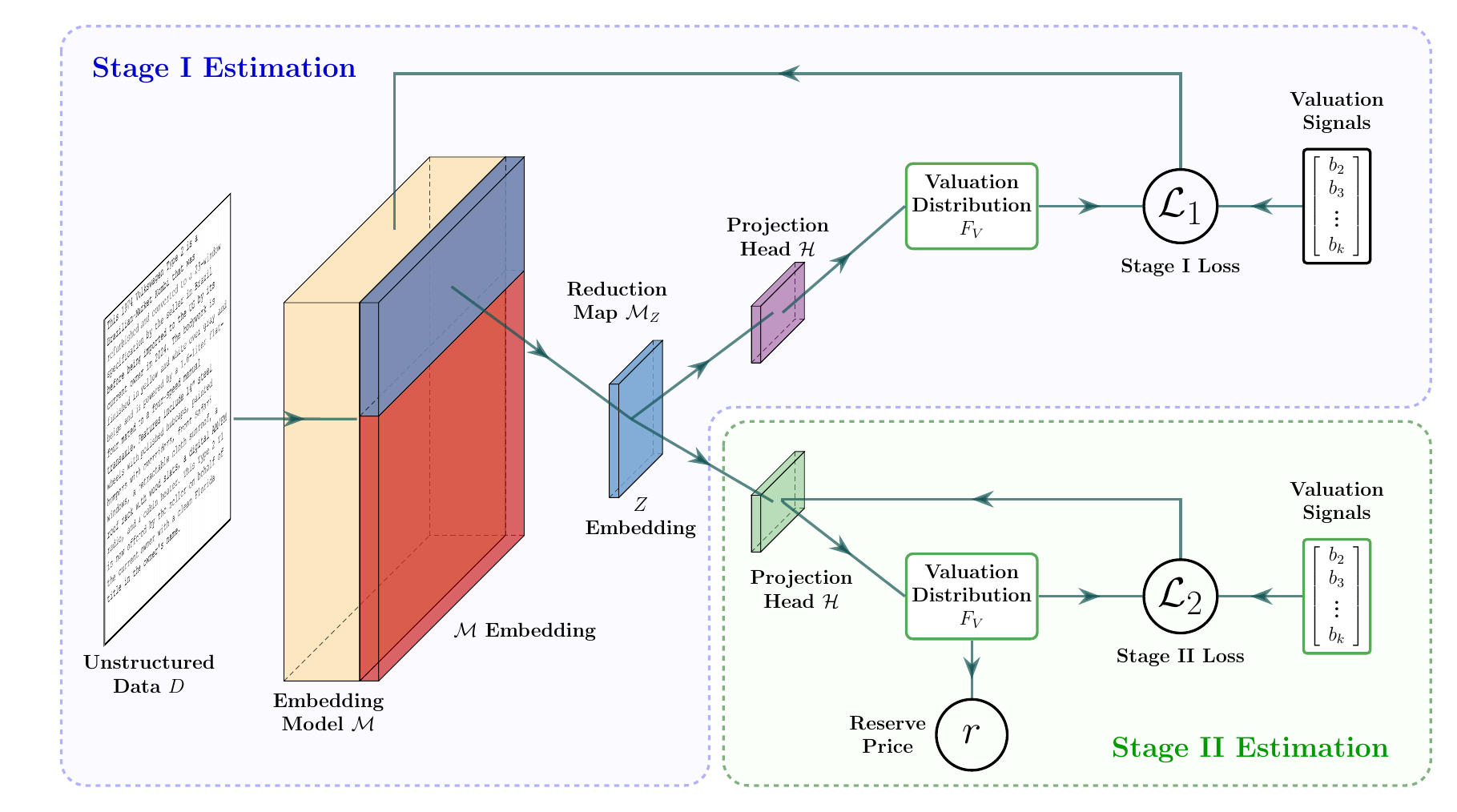}
  \caption{Two Stage Framework for Optimal Reserve Estimation Using Language Model.}
  \label{fig:branch-architecture}
\end{figure}

To address this challenge, we use modern natural language processing methods to estimate the valuation distribution directly from the unstructured text of each auction listing. Our estimator has two stages, each using one of two randomly assigned halves of the sample. Figure \ref{fig:branch-architecture} presents a schematic overview. In the first stage, we fine-tune a text embedding language model to map each listing's text into a low-dimensional representation, and we then freeze the parameters of the language model. The training signal is the structural likelihood of the observed bids, which ties the representation to the valuation distribution that determines the reserve. In the second stage, a projection head maps the fitted language model's low-dimensional embedding to the parameters of a generalized Pareto tail for the listing's valuation distribution, estimated by maximum likelihood on the other half of the sample. The tail has a linear inverse hazard rate, so the optimal reserve follows in closed form from the listing's text and the seller's valuation. We detail this method in Section \ref{sec:model}. 

The central inference problem is that each reserve is evaluated at a representation learned from bids. We address this problem by learning the representation in Stage I and holding the fine-tuned language model fixed during inference in Stage II. Under independence and regularity conditions, the sample reserved for Stage II can be used to estimate the projection head mapping from the learned embedding, and standard sandwich inference gives intervals for its implied reserve \citep{Cox1975Note,White1982Maximum,Newey1994Large}. This argument does not require fine-tuning to converge to a fixed population representation.

To demonstrate the empirical utility of the proposed framework, we collect transactional data from a prominent US online auction platform for used vehicles. Leveraging time-stamped and user-identifiable bids from approximately 65,800 car auctions between 2014 and 2023, we fine-tune a language model to estimate the valuation distribution and subsequently derive optimal reserve prices. In the inference sample, we find lowering reserves that were set above the optimal price would have converted \$\ResPolHighBidTotal{} million in observed final bids into automatic sales. Additionally, we show raising reserves that were set below the optimal price would have lost \$\ResPolLowRevenue{} million in realized revenue because the final bids fell below the model's reserve. The net automatic-sale revenue difference is \$\ResPolNetRevenue{} million which equals \ResPolNetIncumbentShare\% of the \$\ResPolIncumbentRevenueBillion{} billion in observed revenue generated by the incumbent policy.

The principal advantage of our approach lies in the language model's ability to process complex linguistic patterns, contextual details, and nuanced product attributes. Our framework develops this capability into a pricing method through three contributions. First, it connects representation learning to reserve pricing by fine-tuning the language model using the structural auction likelihood. Second, it identifies the valuation-tail information needed for the reserve decision without observing or modeling the number of potential bidders. Third, it separates representation learning from structural estimation, permitting inference conditional on the learned representation. Together, these components reduce reliance on manually specified product attributes while preserving an explicit link between listing text, bidder valuations, and reserve prices. This method is especially valuable in modern digital marketplaces, where product heterogeneity makes manual feature specification challenging and costly \citep{krasnokutskaya2011identification}. Ultimately, this work answers the calls by \cite{kumar2018research,misic2020data,cohen2026om} to integrate artificial intelligence into operations management research and contributes to the burgeoning literature at the intersection of machine learning and structural economic inference \citep{athey2018impact, abbasi2024pathways, sarker2025advancing}.

The rest of the paper proceeds as follows. Section \ref{sec:literature} situates our work within the reserve price literature and the evolving literature connecting natural language processing with structural econometrics. Section \ref{sec:model} develops the model, identification result, and estimation strategy. Section \ref{sec:data} describes the online vehicle auction setting, sample construction, and descriptive evidence. Section \ref{sec:results} presents three applications. These examine optimal reserve prices and the incumbent policy, the effect of listing text on fitted reserves, and the relationship between stated mileage, fitted reserves, and realized second-highest bids. Finally, Section \ref{sec:conclusion} discusses the implications for demand estimation and operational decision making in digital marketplaces.

\section{Literature Review}\label{sec:literature}

This paper joins two research streams. The first is the empirical auction literature on what bids reveal about demand and how bid data can improve auction decisions. The second studies how unstructured information can be used to estimate demand for differentiated products. Their intersection gives rise to the operational problem we study. Occasionally, a platform must set a reserve price for a product that may never have been sold before, without observing how many buyers considered it, while much of the information that distinguishes it from other products appears in its rich description rather than in a fixed set of attributes. 

The empirical auction literature shows how data available to the analyst determine what can be learned about demand. One of the most common methods applies bid order statistics to identify the valuation distribution under symmetric independent private values when bidder counts are observed \citep{athey2002identification}. With weaker restrictions on bidding behavior, the same objective generally yields bounds rather than point identification \citep{Haile2003Inference}. Other studies recover valuations in the presence of common auction heterogeneity, dependence across bidders, or endogenous and unobserved participation by exploiting variation in bids, bidder counts, and reserve prices \citep{krasnokutskaya2011identification,Roberts2013,AradillasLopez2013Identification,Freyberger2022Identification}. Together, these papers clarify which features of demand can be recovered from auction records and which require additional variation or assumptions. This distinction is consequential for reserve setting because the decision depends on the tail of the valuation distribution, not only on a forecast of the final price.

A related branch of the auction literature uses bids to guide platform decisions without first recovering every demand primitive. \citet{HaruvyHejaziNiaOzerSimsek2023} combines early bidding dynamics with an equilibrium adjustment for the winner's curse to forecast terminal bids and inform choices such as minimum bid increments. Work on reserve setting then takes two broad routes. The structural route uses the ranking of upper bids to recover information about valuations. \citet{Song2004Nonparametric}, for example, shows that this information can be recovered without observing the number of potential bidders, a technique we also exploit. Conversely, the direct route learns a pricing rule from auction outcomes, whether by estimating feature-dependent reserves from the top two bids, optimizing reserves without first recovering the valuation distribution, personalizing reserves across bidders, or accounting for bidders who anticipate future reserve updates \citep{MohriMedina2016,allouah2020prior,CoeyLarsenSweeneyWaisman2021,KanoriaNazerzadeh2021,golrezaei2021dynamic,derakhshan2022linear,allouah2022pricing}. These studies show that observed bid histories can support auction policy through either structural recovery or direct optimization. Our identification strategy also begins with bid histories, but the information is carried across auctions differently. Because individual auction objects may appear only once, we pool across differentiated products through their descriptions rather than rely on repeated auctions for comparable goods or bidder-specific histories. Personalization therefore occurs at the listing level, and can set a reserve price for a new listing before it receives any bids.

The second stream examines how firms estimate demand when product-level information is incomplete. Work in operations has confronted this issue when products are new \citep{FerreiraLeeSimchiLevi2016}, assortments change quickly or prices cannot be adjusted often \citep{FerreiraMower2023}, nonpurchases or zero sales conceal part of demand \citep{Pauphilet2024,adam2024machine}, and the set of alternatives each customer considered goes unrecorded \citep{jagabathula2024demand}. Across these settings, the common response is to share information rather than estimate each product in isolation \citep{cohen2020feature,ban2021personalized}. Demand forecasts pool observations across new and previously sold products \citep{hu2019forecasting,cohen2022data}, short learning periods exploit variation across changing assortments, and corrections for missing or selected outcomes recover price sensitivity that observed purchases alone would obscure. This literature therefore treats sparse demand not only as a small-sample problem but also as a question of what information can legitimately travel across products and customers. Our setting has the same informational structure, but ordered bids replace purchase outcomes and the listing description determines how information is shared across differentiated products. The result is a demand estimate for an auction object with no transaction history of its own. 

Unstructured content expands what can be pooled by recovering demand information that conventional product records omit. Research in operations and marketing first demonstrated this value through interpretable measures extracted from language. Text can expose product defects, recover the economic value of product features, and improve demand estimation through the joint information in review sentiment and star ratings \citep{archak2011deriving,abrahams2015integrated,ChoSosaHasija2022}. The latter study is especially close in motivation because it incorporates review text into a choice model of automobile demand. More recent work replaces a small set of analyst-selected textual measures with learned representations. \citet{khern2024selecting} use image representations to improve a platform's selection of restaurant cover images, while \citet{Dew2025} uses text and image representations to measure preferences without requiring a predefined attribute system \citep[see also][]{gabel2022product}. Across these studies, unstructured information becomes operationally useful when the representation preserves variation relevant to a demand or decision model, not merely when it summarizes semantic content.

Recent structural demand research takes this integration further by placing learned product representations inside the economic model. \cite{Compiani2026} use pretrained text and image features in a mixed-logit model to recover substitution patterns among differentiated products. Their approach and ours use learned representations to capture product heterogeneity without requiring a complete set of observed attributes. In their framework, pretrained embeddings are reduced to principal components and incorporated into the demand model, with model selection guided by demand fit. Following the principle of tailoring fine-tuning to the estimation problem emphasized by \cite{vafa2025estimating}, we instead adapt the language model itself using the structural auction likelihood. This allows observed bidding behavior to shape the representation used to estimate the valuation tail and determine the reserve price. This design follows the broader text-as-data principle that a representation should be validated against the economic object it is intended to measure \citep{gentzkow2019text,ash2023text}.

Placing a learned representation inside a structural demand model also changes the interpretation of statistical uncertainty. Orthogonal-score methods address errors in learned nuisance functions when estimating a population target \citep{Chernozhukov2018Double,Farrell2021Deep}. Recent work shows how fine-tuned representations can enter economic decompositions while addressing bias from high-dimensional controls \citep{vafa2025estimating}. We separate representation learning from structural estimation and condition inference on the frozen representation. This provides intervals for the reserve implied by its best fitting valuation model. Taken together, the paper connects a count-free source of auction identification with the operations problem of estimating demand for products that lack their own transaction histories. Unstructured descriptions allow information to travel across products, while the auction model preserves the link between that information, bidder valuations, and the reserve decision.

\section{Model, Identification, and Estimation}\label{sec:model}

We generally apply the following notational conventions. Uppercase letters indicate random variables (or their distributions) with lowercase letters representing their realizations (or their densities). Greek letters denote model parameters, and cursive letters denote model-specific maps. Subscript $j$ indexes the $j^\text{th}$ observed bid or value and its realization, and, when useful, a second subscript $t$ indexes the auction, i.e. $B_{(j),t}$ and $V_{(j),t}$.

\subsection{Valuation Distribution}
\label{sec:distribution}

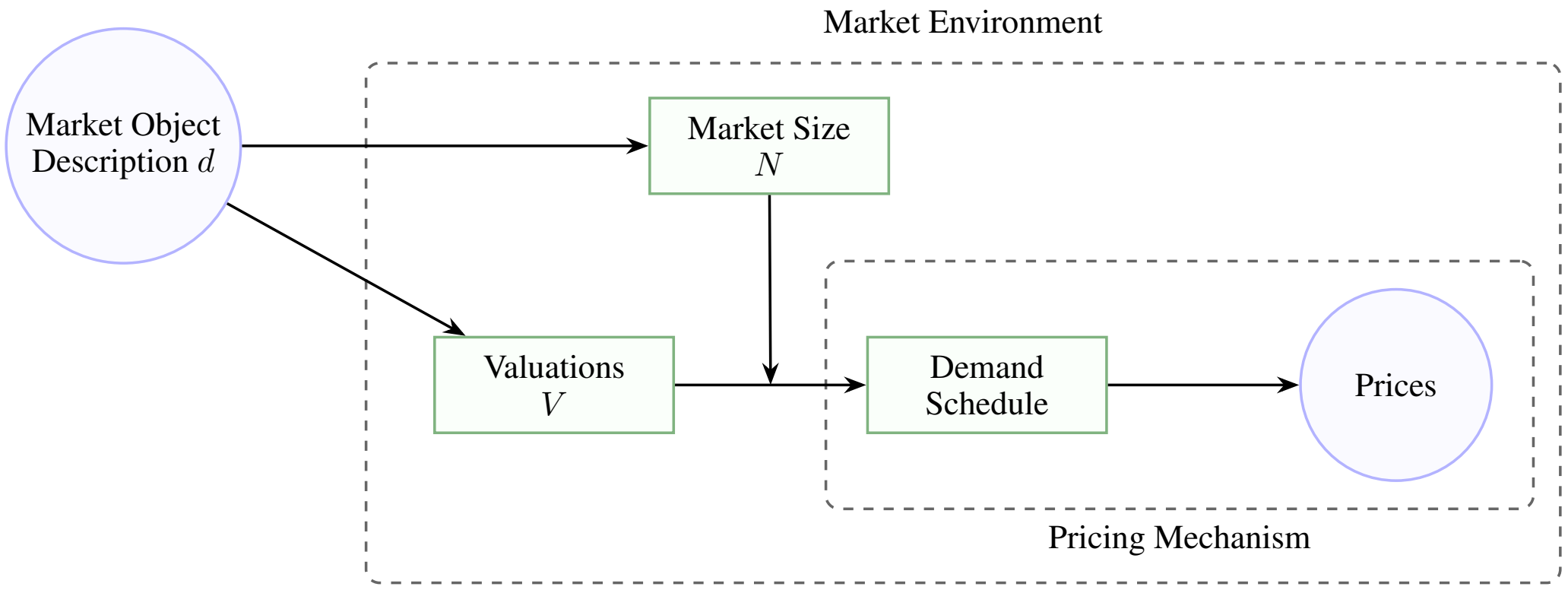
\begin{figure}[t]
  \begin{center}
  \resizebox{0.825\textwidth}{!}{%
    \begin{tikzpicture}[
        concept/.style={
          draw=blue!30, fill=blue!2, thick,
          circle, minimum size=2cm, align=center
        },
        process/.style={
          draw=green!40!black!50, fill=green!2, thick,
          rectangle, minimum width=2.5cm,
          minimum height=1cm, align=center
        },
        arrow/.style={-Stealth, thick},
        group_box/.style={
          draw=black!60, dashed, thick,
          rounded corners, fill=white, fill opacity=0.7
        }
      ]

      \node[concept]                           (D)  {Market Object\\Description $d$};
      \node[process,right=2cm of D, yshift=-2.5cm] (V)  {Valuations\\$V$};
      \node[process,right=4.25cm of D]         (N)  {Market Size\\ $N$};
      \node[process,right=2cm of V]            (DS) {Demand\\Schedule};
      \node[concept,right=2cm of DS]           (P)  {Prices};

      \draw[arrow] (D) -- (V);
      \draw[arrow] (D) -- (N);
      \draw[arrow] (V) -- coordinate[midway] (VDSmid) (DS);
      \draw[arrow] (DS) -- (P);
      \draw[arrow] (N) -- (VDSmid);

      \begin{scope}[on background layer]
        \node[group_box, fit=(V)(N)(P),
          inner xsep=20pt, inner ysep=20pt, yshift=-10pt,
        label={[yshift=5pt]above:Market Environment}] {};
        \node[group_box, fit=(DS)(P),
          inner xsep=12pt, inner ysep=8pt,
        label={[yshift=-1pt]south:Pricing Mechanism}] {};
      \end{scope}

    \end{tikzpicture}
  }
\end{center}
  \caption{Assumed Data Generating Process}
  \begin{footnotesize}
    \textit{Note:} Market observables are in circles, latent objects are in rectangles.
  \end{footnotesize}
  \label{fig:data-flow-1}
\end{figure}

The primary counterfactual exercise of this paper is to vary a seller's reserve while holding demand fixed, which requires a description of how auction data are generated. Figure \ref{fig:data-flow-1} outlines the data-generating process we maintain. We assume the introduction of a product with description $d\in D$ generates the two latent demand primitives shown in rectangles: consumers' valuation (willingness to pay) $V$ and the active market size $N$. Each consumer's valuation is an independent draw from the conditional distribution $F(\cdot \mid d)$, with survival function $S(\cdot \mid d)=1-F(\cdot \mid d)$ and continuous density $f(\cdot \mid d)$ over a support of $[\vlow,\vhigh]$ that is strictly positive on the interior of that support. Given a realized market size $N=n$, a demand schedule is formed generating the valuation order statistics $\Vone \ge \Vtwo \ge \cdots \ge V_{(n)}$. Consumers then engage the auction's pricing mechanism, producing the observed prices. 

Consistent with the related literature, we assume $V$ and $N$ are distributionally independent \citep{LevinSmith1994,athey2002identification,GuerrePerrigneVuong2009,AradillasLopez2013Identification,Roberts2013}. We do not, however, restrict a description $d$ from capturing an underlying shared dependence structure in the construction of $V$ and $N$. In other words, while $V$ and $N$ are assumed conditionally independent, they are not independent unconditionally. Furthermore, we assume that bidders have unitary demand, seeking at most one object in the relevant purchase episode, although they may inspect and bid on several substitute listings while searching. Section \ref{sec:desc_stats} reports bidder-participation and cross-auction overlap diagnostics relevant to the unitary-demand assumption. Critically, the two distribution primitives are invariant to the reserve policy, so recovering them is what permits the reserve policy analysis. 

\subsection{Identification of the Valuation Order Statistic Distribution}
\label{sec:song}

The market in our application is an online English auction, for which bidding up to one's valuation is a weakly dominant strategy \citep{vickrey1961counterspeculation,wolfstetter1996auctions}. We therefore treat the bids of ranks $2,\ldots,k$ as revealing the corresponding valuation order statistics, $B_{(j)}=V_{(j)}$, whereas the highest bid $B_{(1)}$ only bounds $V_{(1)}$ from below and is dropped. We apply a retention rule for the value of $k$ based on the timing of each bidder's final bid for each auction for the consecutive ranks whose bids credibly reveal valuations, which we state explicitly in Section \ref{sec:desc_stats}. 

We now show that the conditional law of consecutive upper order statistics identifies $F(\cdot\mid d)$ up to a multiplicative scale above the retained thresholds. Consequently, the count $N$ never needs to be observed or estimated. Proceeding as in \citet{Song2004Nonparametric}, the result follows from the Markov property of order statistics \citep{Arnold2008First}. 

\begin{lemma}[Count-free conditional density]
  \label{lem:song}
  Let $N\geq k\geq3$, conditional on $V_{(k)}=v_{(k)}$, and $d$, the joint density of the order statistics $(V_{(2)},\ldots,V_{(k-1)})$ on $v_{(2)}>\cdots>v_{(k-1)}>v_{(k)}$ is given by,
  \begin{equation}
    f_{2,\dots,k}(v_{(2)},\ldots,v_{(k-1)}\mid v_{(k)},d)
    =\frac{(k-1)!S(v_{(2)}\mid d)\prod_{j=2}^{k-1}f(v_{(j)}\mid d)}{S(v_{(k)}\mid d)^{k-1}},
    \label{eq:song}
  \end{equation}
  and is zero elsewhere. The density does not depend on the law of $N$, and it uniquely determines the survival ratio $S(v\mid d)/S(v_{(k)}\mid d)$ on $[v_{(k)},\vhigh)$.
\end{lemma}

\noindent\emph{Proof.} See Appendix \ref{app:proofs}.

\medskip

Lemma \ref{lem:song} is central to our identification strategy. The $k=3$ case is due to \citet{Song2004Nonparametric}, and our extension to arbitrary $k$ lets a single auction contribute every rank the retention rule certifies. While alternative routes that recover more of the parent distribution are possible, they do so at the cost of requiring stronger inputs. For example, \citet{AradillasLopez2013Identification} and \citet{Freyberger2022Identification} exploit variation in the number of bidders (which we discard), whereas \citet{Haile2003Inference} drop the requirement that bids reveal valuations and deliver bounds instead. 
Integrating the density of the Lemma \ref{lem:song} over $v_{(2)}>\cdots>v_{(k-1)}>v$ yields,
\begin{equation}
  \Pr\!\left(V_{(k-1)}>v\mid V_{(k)}=v_{(k)},d\right)
  =\left\{\frac{S(v\mid d)}{S(v_{(k)}\mid d)}\right\}^{k-1},
  \qquad v\geq v_{(k)}.
  \label{eq:conditionalsurvivalratio}
\end{equation}
The conditional law in equation \eqref{eq:conditionalsurvivalratio} is invariant to a common multiplicative rescaling of the survival function above the threshold. At a fixed threshold $v_{(k)}$, it therefore identifies the threshold-normalized tail $S(v\mid d)/S(v_{(k)}\mid d)$, which the $(k-1)^\text{th}$ root of equation \eqref{eq:conditionalsurvivalratio} recovers and which is the uniqueness claim of Lemma \ref{lem:song}. 

Additionally, we maintain that each listing reaches far more potential bidders than the few bids it retains. We provide evidence of this claim in Section \ref{sec:desc_stats}, where we show that while the average auction receives approximately a dozen unique bidders it also has well over 700 distinct auction ``watchers''. Therefore, given $N=n$, the threshold $V_{(k)}$ exceeds all but $k-1$ of the $n$ valuations and has expected quantile $1-k/(n+1)$, so when $n$ is large relative to $k$ every retained bid lies in the upper tail of $F(\cdot\mid d)$. We let $\underline{t}(d)$ denote the $\alpha$-quantile of the retained thresholds given $d$, for a small fixed $\alpha$, and assume that retained thresholds above $\underline{t}(d)$ have support $[\underline{t}(d),\vhigh)$. The endpoint $\underline{t}(d)$ is a dollar value that the data reveal, whereas its position in the valuation distribution is not identified from bids alone. Its location in the upper tail is therefore a maintained restriction on the size of the market relative to the retained depth. Identification uses only the conditional densities at thresholds in $[\underline{t}(d),\vhigh)$ and leaves the parent distribution below $\underline{t}(d)$ unspecified. A lower $\alpha$ enlarges the identified region and places its lower endpoint further from the extreme upper tail.

\begin{proposition}[Identification of the reserve policy]
  \label{prop:policyid}
  Fix $d$ with $\int_{\underline{t}(d)}^{\vhigh}S(v\mid d)\,dv<\infty$.
  \begin{enumerate}
    \item[(i)] Two parent laws yield the same conditional densities from equation \eqref{eq:song} at every retained depth $k\geq3$ and threshold $v_{(k)}\in[\underline{t}(d),\vhigh)$ if and only if their survival functions are proportional on $[\underline{t}(d),\vhigh)$. These densities identify the complete distribution only if $\underline{t}(d)=\vlow$.
    \item[(ii)] The inverse hazard $S(v\mid d)/f(v\mid d)$ is identified at every interior $v\geq\underline{t}(d)$.
    \item[(iii)] Let $\tilde F$ be a parent law in the equivalence class of (i) whose virtual valuation $v-\tilde S(v)/\tilde f(v)$ is nondecreasing on $(\vlow,\vhigh)$, and let $\Pa_n(r,\tilde F)$ denote the seller's expected payoff from reserve $r$ with $n$ bidders. Suppose the seller sets the reserve before $N$ is realized and $v_0>\underline{t}(d)-S(\underline{t}(d)\mid d)/f(\underline{t}(d)\mid d)$. The maximizers of $\E_N[\Pa_N(r,\tilde F)-\Pa_N(v_0,\tilde F)]$ over $r\in[v_0,\vhigh)$ lie in $(\underline{t}(d),\vhigh)$ and are common to all such parent laws and all count distributions with $\Pr(N\geq1\mid d)>0$. Any unique interior optimum satisfies $r^*-v_0=S(r^*\mid d)/f(r^*\mid d)$.
  \end{enumerate}
\end{proposition}

\noindent\emph{Proof.} See Appendix \ref{app:proofs}.

\medskip

Proposition \ref{prop:policyid} identifies the survival function above $\underline{t}(d)$ up to a multiplicative scale. The scale is the exceedance probability $S(\underline{t}(d)\mid d)$, which is the share of valuations above the dollar endpoint $\underline{t}(d)$, and remains unidentified whenever $\underline{t}(d)>\vlow$. The scale cancels from the inverse hazard that prices the reserve, so the policy is identified without locating $\underline{t}(d)$ in the valuation distribution. Under regularity on the entire support, the condition \(v_0>\underline t(d)-S(\underline t(d)\mid d)/f(\underline t(d)\mid d)\) ensures that the optimal reserve lies in the identified tail. This condition compares the seller’s value with the virtual valuation at the cutoff, which is below the cutoff itself. Consequently, \(v_0\geq\underline t(d)\) is sufficient, but seller values below \(\underline t(d)\) also satisfy the condition whenever their shortfall is less than the inverse hazard at the cutoff. The reserve policy can therefore be identified even when the seller’s value lies below the region in which we identify the valuation tail. Next, we turn to a tail specification that translates the identified inverse hazard into an optimal reserve rule.

\subsection{Tail Specification and the Reserve Rule}
\label{sec:estimation}
\label{sec:gpd}
\label{sec:map}

Because every retained threshold is an extreme order statistic, the analysis requires the valuation distribution only in its upper tail. For any parent law in the maximum domain of attraction of an extreme value distribution, the excesses over a threshold converge to a generalized Pareto law as the threshold quantile approaches one \citep{Balkema1974Residual,Pickands1975Statistical}. We therefore approximate the distribution above every retained threshold with a generalized Pareto distribution $G_{\xi,\sigma}(\cdot\mid d)$, with shape parameter $\xi$ and scale parameter $\sigma$. The specification covers the identified region $[\underline{t}(d),\vhigh)$ and, for the lowest $\alpha$ share of retained thresholds, extends below $\underline{t}(d)$ to thresholds that the same order-statistic argument places in the upper tail.

In alignment with equation \eqref{eq:conditionalsurvivalratio}, this yields the conditional exceedance distribution, 
\begin{equation}
  \frac{S(v \mid d)}{S(v_{(k)} \mid d)}
  \;=\;
  \left[1 + \xi(d)\frac{v-v_{(k)}}{\sigma(v_{(k)}\mid d)}\right]_{+}^{-1/\xi(d)},
  \qquad v \geq v_{(k)},
  \label{eq:gpd}
\end{equation}
with scale $\sigma(v_{(k)}\mid d)>0$, shape $\xi(d)$, and $[\cdot]_{+}$ denoting the positive part. The case $\xi(d)=0$ is the exponential limit, and a negative shape implies a finite upper endpoint. We require $\sigma(v_{(k)}\mid d)>0$, which makes equation \eqref{eq:gpd} a survival function, and $\xi(d)<1$ to ensure a finite expected value. Consequently, this produces a linear inverse hazard,
\begin{equation}
  \frac{S(v \mid d)}{f(v \mid d)}
  \;=\;
  \sigma(v_{(k)}\mid d)+\xi(d)(v-v_{(k)})
  \;=\;
  \psi(d)+\xi(d)v,
  \qquad v\geq v_{(k)},
  \label{eq:gpdhazard}
\end{equation}
where $\psi(d)\equiv\sigma(v_{(k)}\mid d)-\xi(d)v_{(k)}$ is the intercept. 

To ensure the regularity condition of \citet{myerson1981optimal}, we assume that the virtual valuation $v-S(v\mid d)/f(v\mid d)$ is nondecreasing on the entire support. Above every retained threshold, the virtual valuation implied by equation \eqref{eq:gpdhazard} is linear with slope $1-\xi(d)$, so the restriction $\xi(d)<1$ verifies regularity in the upper tail and the reserve requires no ironing. Substituting equation \eqref{eq:gpdhazard} into the reserve rule equation \eqref{eq:myerson1} yields,
\begin{equation}
  r^{*}(v_0\mid d)
  \;=\;
  \frac{v_0+\psi(d)}{1-\xi(d)}.
  \label{eq:gpdreserve}
\end{equation}
Under this specification, the condition of Proposition \ref{prop:policyid}(iii) reads $v_0>(1-\xi(d))\underline{t}(d)-\psi(d)$, which is equivalent to $r^{*}(v_0\mid d)>\underline{t}(d)$. The rule in equation \eqref{eq:gpdreserve} therefore applies on the domain $r^{*}(v_0\mid d)>\underline{t}(d)$, provided the reserve also lies in the support implied by $(\psi(d),\xi(d))$, and the fitted tail verifies the condition listing by listing. 

\subsection{The Description Embedding}
\label{sec:index}
The above model formulation has relied on the underlying premise that the text description $d$ can be incorporated for statistical inference about the underlying valuation distribution of an auctioned item. Next we describe explicitly our proposed method using natural language processing models. Figure \ref{fig:branch-architecture} presented in Section \ref{sec:introduction} summarizes the architecture of the proposed two-stage estimation method, designed specifically to reflect our assumed data generation process. 

In Stage I, we fine-tune a pre-trained language model against the structural likelihood on one half of the sample and then freeze the result \citep{Hu2022Lora,vafa2025estimating}. Conditioning on the frozen language model, Stage II uses the remaining half of the sample to estimate a generalized Pareto tail for the valuation distribution and derive the closed-form reserve rule \citep{Balkema1974Residual,Pickands1975Statistical}. This rule prices any listing from its description $d$ (via the condensed index $z$) and a declared seller value $v_0$. Splitting the sample between the two stages yields parametric inference conditional on the fine-tuned language model \citep{Chernozhukov2018Double,Newey1994Large}. The remainder of the section develops each component in turn.

We formalize the embedding language model $\mathcal{M}$ as a measurable mapping that transforms each description into a numerical representation \citep{Bengio2000Neural,Mikolov2013Linguistic},
\begin{equation*}
  \mathcal{M}(d)\in\mathbb{R}^{q}.
\end{equation*}
In our application, the specific text embedding model $\mathcal{M}$ is Qwen3-Embedding-0.6B \citep{Zhang2025Qwen3}, a text embedding checkpoint obtained by contrastively adapting the Qwen3 base language model \citep{Yang2025Qwen3}, with 28 layers and hidden width $q=1{,}024$. 

The curse of dimensionality makes flexible structural estimation on the language model's full representation infeasible. We therefore compose the embedding model with the fixed reduction map $\mathcal{M}_Z$ of Figure \ref{fig:branch-architecture} to provide the description's embedding,
\begin{equation}
  z(d)=(\mathcal{M}_Z\circ\mathcal{M})(d)\in\mathcal{Z}\subset\mathbb{R}^{p},
  \qquad p\ll q.
  \label{eq:index}
\end{equation}
In our application, the reduction retains the leading $p=64$ coordinates of the pooled final-token state and normalizes them to unit length. A simple truncation function is admissible because this model was trained under Matryoshka representation learning \citep{Kusupati2022Matryoshka}, which makes every leading prefix a trained embedding in its own right. We treat the reduced dimension embedding $z$ as a coarse but sufficient representation of the raw description in the spirit of \cite{Li1991} and \cite{Cook2007}. To ensure that the reduction preserves the relevant economic heterogeneity, we impose the following condition.
\begin{assumption}[Description-embedding sufficiency]
  \label{ass:sufficiency}
  There exists a conditional distribution $\widetilde{F}\bigl(\cdot\mid z(d)\bigr)$, such that for every $d\in D$, and every $v$ in the support of $F(\cdot\mid d)$,  
  \begin{equation}
    \widetilde{F}\bigl(v\mid z(d)\bigr) = F(v\mid d)
  \end{equation}
\end{assumption}

Assumption \ref{ass:sufficiency} requires the description's embedding to preserve the information in listing text relevant to the valuation distribution. This is a maintained restriction on the representation, rather than a consequence of dimensionality reduction or pretraining. Structural fine-tuning uses observed bids to direct the representation toward the valuation features entering the reserve decision, but does not guarantee sufficiency. We show fine-tuning improves held-out dollar prediction, with likelihood comparisons and support diagnostics reported in Appendix \ref{sec:head-to-head}. 

\subsection{Two-Stage Estimation and Conditional Inference}
\label{sec:training}
\label{sec:asymptotics}

Our identification method in equation \eqref{eq:song} relies on auction $t$ contributing $k_t-1$ valuation-revealing bids, ordered $B_{(2),t}>\cdots>B_{(k_t),t}$, to estimate the valuation distribution of the object with description $d_t$. Substituting equation \eqref{eq:gpd} into equation \eqref{eq:song} and discarding the combinatorial constant gives the log-likelihood contribution of auction $t$ at tail parameters $(\psi_t,\xi_t)$,
\begin{equation}
  \begin{aligned}
  \ell_t(\psi_t,\xi_t)
  &=
  -(k_t-2)\log\sigma_t
  -\sum_{j=2}^{k_t-1}
  \left(\frac{1+\one\{j=2\}}{\xi_t}+1\right)
  \log\left[1+\xi_t\,\frac{B_{(j),t}-B_{(k_t),t}}{\sigma_t}\right], \\
  \sigma_t&=\psi_t+\xi_t B_{(k_t),t}.
  \end{aligned}
  \label{eq:gpdloglik}
\end{equation}
The contribution involves neither $N$ nor the exceedance probability $S(B_{(k_t),t}\mid d_t)$, and every retained auction enters it, so nothing estimated below depends on the number of potential bidders.

To connect a description embedding $z(d)$ with the generalized Pareto distribution parameters  $(\psi,\xi)$, we append a single-layer fully connected network head $\mathcal{H}$ to the language model. We parameterize the head $\mathcal{H}$ with links that enforce the parameter space (i.e. 64$\times$2). After coordinate-wise standardization over the fitting sample, $z_{\mathrm{std}}=\operatorname{diag}(\tau)^{-1}(z-\mu)$, the tail parameters are given by
\begin{align*}
  \xi(z)&=\underline{\xi}+(\bar{\xi}-\underline{\xi})\Lambda\!\left(\delta+\beta z_{\mathrm{std}}\right),\\
  \sigma(\tilde{b}\mid z)&=\underline{\sigma}+\log\!\left(1+e^{\eta+\rho z_{\mathrm{std}}}\right),\\
  \psi(z)&=\sigma(\tilde{b}\mid z)-\xi(z)\,\tilde{b},
\end{align*}
where $(\mu,\tau)$ are the coordinate-wise mean and standard deviation of $z(d_t)$; $\underline{\xi}=-0.45$ and $\bar{\xi}=1.1$ are fixed lower and upper bounds on the shape parameter. $\Lambda$ is the logistic function, $\underline{\sigma}=\$1$ is a positive floor on the scale at the reference threshold, $\log(1+e^{x})$ is the \emph{softplus} function, and $\tilde{b}=\$10{,}000$ is a reference threshold. We estimate on the dollar scale, in units of \$10,000, because the generalized Pareto family is closed under affine maps, whereas the linear inverse hazard in equation \eqref{eq:gpdhazard} does not survive a logarithmic transformation.

The head coefficients $\gamma=(\beta,\delta,\rho,\eta)$ collect the slopes $\beta,\rho\in\mathbb{R}^{1\times p}$ and the intercepts $\delta,\eta\in\mathbb{R}$, giving $2(p+1)=130$ parameters. Together with the standardization constants, they form $\vartheta=(\mu,\tau,\gamma)$. We write $h_{\vartheta}(z)=(\psi_{\vartheta}(z),\xi_{\vartheta}(z))$ for the resulting head. The sample criterion for a set of $T$ auctions with the associated embedding matrix $z$ is
\begin{equation*}
  \mathcal{L}(\gamma,z;T)
  \;=\;
  -\frac{1}{|T|}\sum_{t\in T}\ell_t\bigl(h_{\vartheta}(z(d_t))\bigr),
\end{equation*}
 
The two stages of Figure \ref{fig:branch-architecture} minimize this criterion on randomly assigned, disjoint sample halves: $T_{\mathrm{I}}$ for representation learning and $T_{\mathrm{II}}$ for inference. Stage I fine-tunes the representation using the economic likelihood, following the principle of tailoring training to the estimation problem emphasized by \citet{vafa2025estimating}. Stage II takes the frozen fine-tuned language model and re-estimates the head on independent auctions. This separation draws on the sample-splitting logic of \citet{Chernozhukov2018Double} and inference for targets defined by a selected model \citep{Cox1975Note,Rinaldo2019Bootstrapping}. Conditional on the frozen fine-tuned language model, Stage II is a stacked two-step estimator \citep[Section 6]{Newey1994Large}, yielding parametric inference for the associated head and reserve under the regularity conditions we state in Appendix \ref{sec:conditional-target}.

Stage I augments the embedding model with a low-rank adapter on its attention projections \citep{Hu2022Lora}, with adapter parameters $\theta$ normalized so that $\theta=0$ returns the pre-trained embedding from equation \eqref{eq:index}, and solves
\begin{equation*}
  (\hat{\theta},\hat{\gamma}_{\mathrm{I}})
  \;\in\;
  \argmin_{\theta,\,\gamma}\mathcal{L}(\gamma,z_{\theta};T_{\mathrm{I}}),
\end{equation*}
selecting the adapter rank, learning rate, stopping epoch, and initialization by the likelihood on a validation slice held out within $T_{\mathrm{I}}$. The resulting adapter parameters $\hat{\theta}$ are then frozen, and the Stage I head $\hat{\gamma}_{\mathrm{I}}$ serves only as one starting value for Stage II, which treats $\hat{z}$ as the fixed transformation parameters of text and re-estimates the head on the inference half,
\begin{equation}
  \hat{\gamma}
  \;\in\;
  \argmin_{\gamma}\mathcal{L}(\gamma,\hat{z};T_{\mathrm{II}}).
  \label{eq:stage2}
\end{equation}
We write $\hat{\vartheta}=(\hat{\mu},\hat{\tau},\hat{\gamma})$ for the Stage II estimate together with its standardization constants. Equivalently, $\hat{\vartheta}$ solves a stacked system of estimating equations, namely the moment conditions that define $(\hat{\mu},\hat{\tau})$ on $T_{\mathrm{II}}$ and the first-order conditions of equation \eqref{eq:stage2}.

Holding Stage I fixed, let $\vartheta^{\dagger}$ collect the (true) population standardization constants and the head coefficients, and let $r^{\dagger}(v_0\mid d)$ be the reserve that equation \eqref{eq:gpdreserve} assigns to the tail parameters $h_{\vartheta^{\dagger}}(\hat z(d))$. Each realized representation defines its own target, so the inference requires no assumption that fine-tuning converges to a population representation. Conditional on the Stage I run, the sandwich variance of the stacked estimator and the delta method deliver standard errors and confidence intervals for $r^{\dagger}(v_0\mid d)$ that remain valid for this target when the tail or the head is misspecified \citep{Huber1967Behavior,White1982Maximum}. Appendix \ref{sec:conditional-target} states the conditioning set, the estimating equations, and the regularity conditions. Under the maintained auction assumptions, correct specification, and Assumption \ref{ass:sufficiency} for $\hat z$, the target equals the optimal reserve on the domain of equation \eqref{eq:gpdreserve}.

\section{Data: Online Vehicle Auctions}\label{sec:data}

\subsection{The Auction Environment}

We collected transactional data from \href{https://bringatrailer.com}{bringatrailer.com} (BaT), one of the largest online vehicle auction marketplaces in the United States, covering every auction the platform held from July 2014 through December 2023. BaT runs an ascending English auction with a secret reserve price $r\ge 0$. The description of each auction vehicle is written by BaT staff following a stylized template that reviews the history of the vehicle, its current condition, any modifications or restorations, and its provenance. For roughly 24\% of auctions in our dataset, BaT revealed that the auction carried no reserve by labeling it ``No Reserve'' on the auction page.

Each auction begins at \$1 (all prices in USD), typically on a seven-day clock. To discourage snipe bidding (cf. \citealt{roth2002last}), BaT extends the clock by two minutes whenever a new bid arrives within two minutes of the scheduled close. When the auction concludes, BaT privately compares the highest bid $b_{(1)}$ to the vehicle's secret reserve price $r$ and automatically announces whether the reserve was met. If the reserve is met ($b_{(1)}\geq r$), the winning bidder pays the seller the amount $b_{(1)}$. During the period of data collection, BaT charged the winning bidder a transaction fee of 5\% of $b_{(1)}$, subject to a minimum of \$250 and a maximum of \$5,000, that is, $\min(\max(0.05\times b_{(1)}, \$250), \$5{,}000)$. If the reserve is not met ($b_{(1)}<r$), BaT connects the highest bidder with the seller to attempt a privately negotiated sale. Negotiated outcomes are unobserved, and our reserve rule assumes sellers offer exactly the reserve and high bidders accept whenever their valuations permit. With weaker seller bargaining power, the rule provides an upper bound within the continuation model detailed in Appendix \ref{app:mechanism}. Regardless of the outcome, the seller pays an unrecoverable listing fee of \$99.

All bids and text-based communications submitted during an auction are publicly recorded in a chat section at the bottom of the auction's webpage. Every submitted bid is therefore timestamped and attributed to a persistent, anonymous bidder identifier. We obtained all market data from these chat sections.

\subsection{Sample Selection Criteria and Descriptive Statistics}\label{sec:desc_stats}

Our identification strategy from Lemma \ref{lem:song} requires direct observation of the top $k$ valuation ranks drawn from $V$, and the sample must be tailored to that requirement. In the English auction equilibrium associated with BaT's mechanism, bidders have an incentive to bid up to their private valuations, since they can avoid bidding more than they are willing to pay and will not let another bidder win at a price they are willing to pay \citep{Haile2003Inference}. Valuation revelation can generally be assumed for the price-setting bid, so that $B_{(2)}=V_{(2)}$. Extending the property to all ranks $2,\dots,k$ requires justification, because bid increments, jump bidding, and strategic delay can leave observed bids as bounds on valuations rather than as the valuations themselves \citep{Bapna2009Overlapping}.

We implement a retention rule through selection criteria that isolate the auctions and the bid ranks for which a bidder's terminal action is a credible proxy for that bidder's valuation. For our dataset, we let the retention window open five hours before the rule-defined close. The rule imposes two conditions. The timing condition requires that the bidder's terminal action, the last bid that bidder submits in the auction, fall inside the window. The threshold-crossing condition of \citet{Song2004Nonparametric} requires, for every rank $j\geq 3$, that the lowest cutoff standing among the eventual top two bidders when the window opened lie strictly below $B_{(j)}$. Retention stops at the first rank that fails either condition, so every retained auction contributes a consecutive prefix $B_{(1)},\dots,B_{(k)}$, and an auction meets the bid-retention criteria when $k\geq 3$.

Of the 89,080 auctions BaT held over the sample period, 65,788 auctions, or approximately 74\% of all auctions held, met our selection criteria. These auctions provide 256,186 retained rank contributions, with a maximum retained depth of $k=22$. Table \ref{tab:summary-1} describes this selected sample. Among the 65,788 auctions, $|T_{\mathrm{I}}|=32{,}902$ listings are used for representation learning and $|T_{\mathrm{II}}|=32{,}886$ used for structural estimation and inference. Once selected, we construct each model input using a fixed text template that combines the vehicle title, auction closing date and time, listing location, seller type (dealership or private owner), and platform-written vehicle description. These fields jointly constitute the description $d$ supplied to the text embedding model. 

Among our selected auctions, bidding is heavily concentrated at the close. Among the terminal bids of all bidders, 54.2\% are submitted on the final day, 37.5\% within the last hour, 27.0\% in the last 15 minutes, and 16.9\% in the final five minutes. The concentration is far sharper for the ranks that enter the likelihood, with 98.0\% of second-highest and 91.0\% of third-highest bidders acting within the final hour. The selected sample engages 190,140 unique bidders, each participating in an average of 4.469 auctions. Despite this active engagement, 77.4\% of these bidders never won an auction, only 4.9\% won at least twice, and 0.7\% won five or more auctions.

\begin{table}[htbp]
  \centering
  \caption{Descriptive Statistics of Selected BringATrailer.com Auctions (July 2014--December 2023)}
    \begin{tabular}{lcccc}
    \hline\hline
    \textbf{65,788 Auctions} & \textbf{Mean} & \textbf{St. Dev.} & \textbf{Min.} & \textbf{Max.} \\
    \hline \\ [-2.25ex]
    Bidding Duration (Days) & 7.002 & .626 & .002 & 17.701 \\
    $\one[\text{Sold by Dealer}]$ & .314 & .464 & 0 & 1 \\
    $\one[\max(b)\geq r]$ (Reserve Met) & .787 & .409 & 0 & 1 \\
    No. of Active Bidders & 12.916 & 4.230 & 3 & 39 \\
    Auction Views (1000s) & 14.236 & 9.430 & .001 & 358.191 \\
    Auction Watchers (1000s) & .746 & .346 & .074 & 6.105 \\
    \hline \\ [-2.25ex]
    \multicolumn{1}{l}{\textbf{Nominal Bid Values}} &  &  &  &  \\\cline{1-1}
    \\ [-2.25ex]
    Winning Bid (in 1000s) & \$44.471 & \$72.168 & \$.500 & \$5,360 \\
    \\ [0.75ex]
    \multicolumn{1}{l}{\textbf{Normalized Terminal Bids}} &  &  &  &  \\\cline{1-1}
    \\ [-2.25ex]
    1$^\text{st}$ Highest Bidder (Winner) & 1 & -- & 1 & 1 \\
    2$^\text{nd}$ Highest Bidder & .975 & .030 & .370 & 1.000 \\
    3$^\text{rd}$ Highest Bidder & .896 & .089 & .088 & .998 \\
    4$^\text{th}$ Highest Bidder & .814 & .120 & .001 & .994 \\
    All Bidders & .614 & .291 & .000 & 1 \\
    \hline \\ [-2.25ex]
    \textbf{Fraction of Terminal Bids} &  &  &  &  \\
    \textbf{Submitted in Final...} & \textbf{5 Min.} & \textbf{15 Min.} & \textbf{1 Hr.} & \textbf{24 Hr.} \\
    \hline \\ [-2.25ex]
    1$^\text{st}$ Highest Bidder (Winner) & .945 & .977 & .995 & 1 \\
    2$^\text{nd}$ Highest Bidder & .874 & .932 & .980 & 1 \\
    3$^\text{rd}$ Highest Bidder & .295 & .745 & .910 & 1 \\
    4$^\text{th}$ Highest Bidder & .055 & .452 & .711 & .923 \\
    All Bidders & .169 & .270 & .375 & .542 \\
    \hline\hline \\ [-2.25ex]
    \textbf{190,140 Unique Bidders} & \textbf{Mean} & \textbf{St. Dev.} & \textbf{Min.} & \textbf{Max.} \\
    \hline \\ [-2.25ex]
    No. Auctions Participated & 4.469 & 13.952 & 1 & 1,758 \\
    \hline \\ [-2.25ex]
    \textbf{Fraction of Bidders} &  &  &  &  \\
    \textbf{Who Won...Auctions} & \textbf{0} & \textbf{$\geq$1} & \textbf{$\geq$2} & \textbf{$\geq$5} \\
    \hline \\ [-2.25ex]
    & .774 & .226 & .049 & .007 \\
    \hline \\ [-1.5ex]
    \multicolumn{5}{l}{Selected sample satisfies the five-hour retention rule. A bidder's}\\
    \multicolumn{5}{l}{terminal bid is their last action. Ranks are over bidders, not bid rows.}\\
  \end{tabular}

  \label{tab:summary-1}
\end{table}

Over the included listings, approximately 31.4\% of the auctions were managed by car dealers rather than private owners, and the reserve was met in 78.7\% of cases. The observed count of unique bidder identifiers averaged 12.916 per auction, with a minimum of 3 and a maximum of 39. The average winning bid across the auctions was \$44,471, with the highest recorded bid reaching \$5,360,000 and the lowest at \$500. Normalizing each auction by its winning bid, the terminal bids of the second, third, and fourth highest bidders average 97.5\%, 89.6\%, and 81.4\% of the winning bid respectively. Throughout, a rank refers to a bidder's terminal bid rather than to an individual bid row, since the former is the object at which the likelihood of \eqref{eq:gpdloglik} is evaluated. Ranking bid rows instead would compress the reported spacings, because consecutive rows late in an auction are often successive bids by the same pair of bidders. 

The retained bids are a small fraction of each listing's audience. Among the 63,643 selected auctions that report audience counts, recorded after the close, the median listing has 12,015 page views and 681 watchers, against a median retained depth of five. The observed bidder count understates the market size $N$ of Section \ref{sec:distribution}, because $N$ includes every consumer who draws a valuation, among them those whose valuations never exceeded the standing price and who therefore never bid. If every watcher is among the consumers who draw a valuation, so that $N$ is at least the watcher count, the median threshold lies at an expected quantile of at least 0.992 of the valuation distribution, and at least 0.984 for nine in ten listings. Every retained bid is therefore an extreme order statistic in the sense of Section \ref{sec:song}.

\begin{figure}[htbp]
  \centering
  \includegraphics[width=0.90\textwidth]{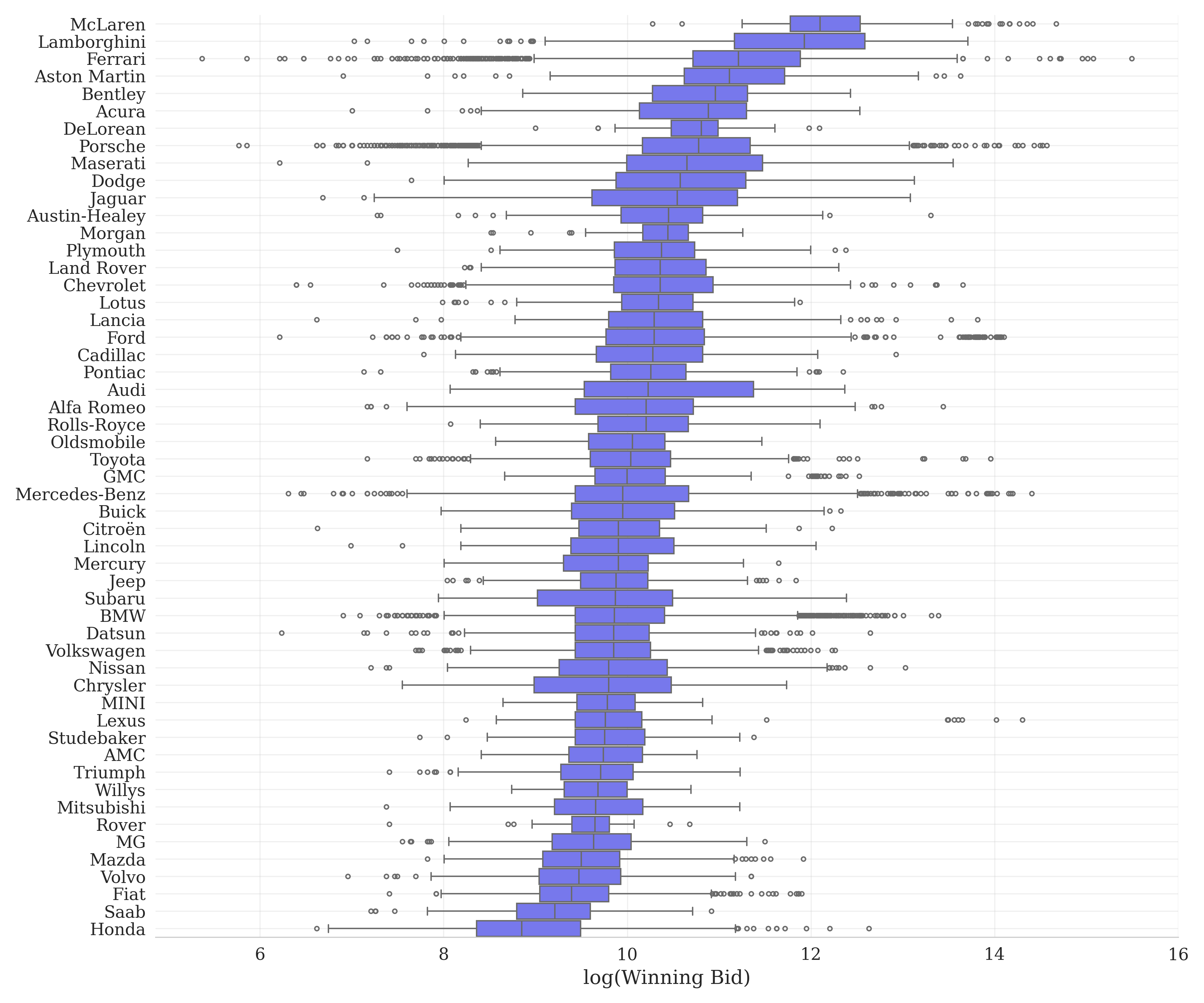}
  \caption{Distribution of the Winning Bid (Log-Transformed) for Vehicle Makes with More than 50 Listings}
  \label{fig:distribution-winningbid}
\end{figure}

Figure \ref{fig:distribution-winningbid} displays the distribution of the winning bid (log-transformed) for the car makes with more than 50 auction listings. Winning bids vary widely within every make. Common structured factors such as model, year, engine, and mileage explain part of this heterogeneity, and the qualitative attributes described in the listing text plausibly carry much of the remainder. The fine-tuned language model of Section \ref{sec:model} is constructed to absorb that residual heterogeneity.

Finally, we examine the extent to which overlapping auctions, whose outcomes may be interdependent, could undermine our identification. To do so, we use bidder identifiers and auction closing times to measure cross-auction overlap. Because BaT auctions typically remain live for about seven days, a seven-day closing window provides a broad proxy for auctions that could have overlapped while live, while 24-hour and 6-hour windows speak more directly to the late-bidding period most relevant for our identification strategy. On the participation margin, repeat bidding is generally not tightly clustered in time. Among bidders who participate in more than one auction, the median gap between successive auction closings is 11.96 days when all adjacent gaps are pooled across repeat bidders and 39.01 days when each bidder is first summarized by their own median adjacent gap. On the more relevant price-setting margin, 66.9\% of auctions have no top-three bidder who is also top-three in another auction closing within seven days, and only about 5.0\% have two or more such overlaps. Within 24 hours, only 10.2\% of auctions have any such overlap and only 0.47\% have two or more. Within 6 hours, the corresponding figures are 6.1\% and 0.22\%. These diagnostics count overlap with any nearby auction rather than only with close substitutes, so they likely overstate the scope for economically meaningful cross-auction substitution. Taken together, the evidence suggests that cross-auction overlap is present but limited for the typical auction, and especially limited among the price-setting bidders in the short windows closest to the close.

\section{Results}\label{sec:results}

We report the results in three parts. In Section \ref{sec:reserve-estimates}, we report reserve prices for the inference-half listings at calibrated seller values and audit the incumbent reserve policy against our model's reserve. In Section \ref{sec:token-attribution}, we explore the ability of our text-based analysis to identify the words in the description that move the reserve. In Section \ref{sec:mileage}, we examine the mileage gradient of the reserve.

\subsection{Optimal Reserves and the Incumbent Policy}\label{sec:reserve-estimates}

Our main result compares the auction house's incumbent reserve policy with the model's optimal reserve policy while preserving no-reserve outcomes. In \ResPolNHigh{} auctions, the final bid exceeded the model's reserve but did not meet the seller's reserve. Lowering those reserves would have converted \$\ResPolHighBidTotal{} million in observed final bids into automatic sales under reserve-invariant bidding. Applying the model's higher reserve to \ResPolNLow{} successful reserve auctions would instead have placed \$\ResPolLowRevenue{} million in realized revenue at risk because their final bids fell below the model's reserve. The net automatic-sale revenue difference is \$\ResPolNetRevenue{} million. This difference equals \ResPolNetIncumbentShare\% of the \$\ResPolIncumbentRevenueBillion{} billion in observed revenue generated by the incumbent policy. After an unmet reserve, the platform connects the seller and the high bidder for a private negotiation whose outcome and price the data do not record, so the revenue comparison covers only automatic sales at the auction close and does not identify the effect on total transaction revenue or seller payoff. Next we explain our seller value calibration process, then report the reserves and their precision, and finally examine these disagreements with the incumbent policy.

In our analysis we treat the seller value $v_0$ as a policy input, and therefore must calibrate it. To do this, we set $v_{0}=c\cdot\widehat{\E}[B_{(2)}\mid\hat{z}]$, where the anchor is a ridge regression of the log second-highest bid on the frozen fine-tuned language model, fit on $T_{\mathrm{I}}$ and applied out of sample to $T_{\mathrm{II}}$, and $c$ is a multiplier that we calibrate. The calibrated multiplier $c^\ast$ matches the incumbent sale rate of \ResMetTargetOne\% among the \ResPolicyEligible{} reserve listings in $T_{\mathrm{II}}$, while the \ResPolicyNoReserve{} no-reserve outcomes remain unchanged. The calibration therefore uses one $T_{\mathrm{II}}$ outcome, the sale rate among reserve listings, and every interval we report below is conditional on $c^\ast$. Panel B of Table \ref{tab:reserves} demonstrates how varying $c$ affects the reserve-met share, showing the schedule is steep and monotone. Specifically, the median reserve rises from \$\ResScheduleMedianCzero{} at $c=0$, a reserve that \ResScheduleMetCzero\% of winning bids meet, to \$\ResScheduleMedianCone{} at $c=1$, which only \ResScheduleMetCone\% meet. A reported reserve is therefore meaningful only together with its multiplier, and we state the multiplier wherever we report one.

Panel A of Table \ref{tab:reserves} reports the estimated reserves. Two patterns stand out. First, the reserves are dispersed across listings and precise within them. At $c^\ast=\ResCStar$ the median reserve is \$\ResMedianReserve, with an interquartile range of \$\ResReserveQlo{} to \$\ResReserveQhi, so reserves vary widely across listings at a common multiplier. The median 95\% sampling interval is \$\ResMedianWidth{} wide, \ResMedianWidthShare\% of the reserve, with delta-method sandwich standard errors conditional on the frozen fine-tuned language model and on $c^\ast$. Second, the tail is heavy for nearly every listing. The fitted shape $\xi$ is positive in \ResXiPositiveShare\% of listings and below one everywhere, so the reserve rule \eqref{eq:gpdreserve} applies without ironing. A positive shape means the inverse hazard $\psi(z)+\xi(z)v$ rises with the valuation, so the density of valuations at $v$ relative to the mass above it falls as $v$ rises. The marginal cost of holding out for a higher bid therefore shrinks as the tail grows heavier, and the optimal reserve rises with $\xi$.

Proposition \ref{prop:policyid}(iii) prices a listing whenever its optimal reserve exceeds $\underline{t}(d)$, a condition the fitted tail verifies listing by listing. We set $\alpha=0.05$ and estimate $\underline{t}(d)$ for each listing by the 5th percentile of the retained thresholds of its 512 nearest $T_{\mathrm{II}}$ listings from the fine-tuned language model, with the 95th percentile as an upper reference. The reserve lies inside this band for 32,845 of the 32,886 listings, with \ResSupportBelow{} below and \ResSupportAbove{} above. The seller value itself lies above $\underline{t}(d)$ for \ResSupportVzeroShare\% of listings, so most listings satisfy the condition of Proposition \ref{prop:policyid}(iii) with margin. The choice of $\alpha$ trades the size of the identified region against the depth in the tail at which it begins. Raising $\alpha$ to 0.25, with the 75th percentile as the upper reference, moves \ResBandNarrowBelow{} reserves below the band and \ResBandNarrowAbove{} above it, whereas the neighbor count matters little (Appendix \ref{app:computation}). We therefore read the support check as evidence that the reserves lie in the region the data inform.

\begin{table}[!htbp]
  \centering
  \caption{Optimal Reserves at the Calibrated Seller Value}
  \label{tab:reserves}
\begin{tabular}{lr}
\toprule
\multicolumn{2}{l}{\emph{Panel A. Reserves at the calibrated seller value}} \\
\midrule
Reserve, 25th percentile (\$) & 12,978 \\
Reserve, median (\$) & 20,686 \\
Reserve, 75th percentile (\$) & 33,947 \\
Tail shape $\xi$, 25th percentile & 0.105 \\
Tail shape $\xi$, median & 0.143 \\
Tail shape $\xi$, 75th percentile & 0.176 \\
Share of listings with $\xi>0$ (\%) & 99.96 \\
Median standard error (\$) & 177 \\
Median 95\% interval width (\$) & 694 \\
Median interval width, share of reserve (\%) & 3.5 \\
Reserves inside the threshold support band & 32,812 \\
Reserves below the band & 38 \\
Reserves above the band & 3 \\
Met share at $c^\ast$ (\%) & 72.30 \\
Empirical reserve-met rate (\%) & 72.30 \\
Anchor out-of-sample log $R^2$ & 0.634 \\
Calibrated multiplier $c^\ast$ & 0.467 \\
\midrule
\multicolumn{2}{l}{\emph{Panel B. Reserve schedule in the seller-value multiplier $c$}} \\
\end{tabular}

\begin{tabular}{lrrrr}
Multiplier $c$ & Median (\$) & 25th percentile (\$) & 75th percentile (\$) & Met share (\%) \\
\midrule
0.0 & 3,808 & 2,664 & 5,349 & 99.9 \\
0.2 & 11,054 & 7,122 & 17,570 & 96.2 \\
0.467$^\ast$ & 20,686 & 12,978 & 33,947 & 72.3 \\
0.6 & 25,475 & 15,877 & 42,072 & 56.7 \\
0.8 & 32,694 & 20,242 & 54,292 & 37.2 \\
1.0 & 39,920 & 24,602 & 66,503 & 24.0 \\
\bottomrule
\end{tabular}

  \par\smallskip
  \begin{minipage}{0.92\textwidth}
    \footnotesize \textit{Note.} Panel A reports the 32, 886inference-half listings under the learned Stage I index with its Stage II fit, seller-value anchor, and calibrated multiplier $c^\ast$. Standard errors are delta-method sandwich standard errors conditional on the frozen index and on $c^\ast$, with auctions treated as independent. The support band is the 5th to 95th percentile of the retained thresholds of the 512 nearest inference-half listings in the index. Panel B varies the seller-value multiplier $c$ under the fitted model. The met share is calculated among reserve listings only. No-reserve outcomes remain unchanged in the policy comparison. The starred row is $c^\ast$.
  \end{minipage}
\end{table}

\begin{figure}[!htbp]
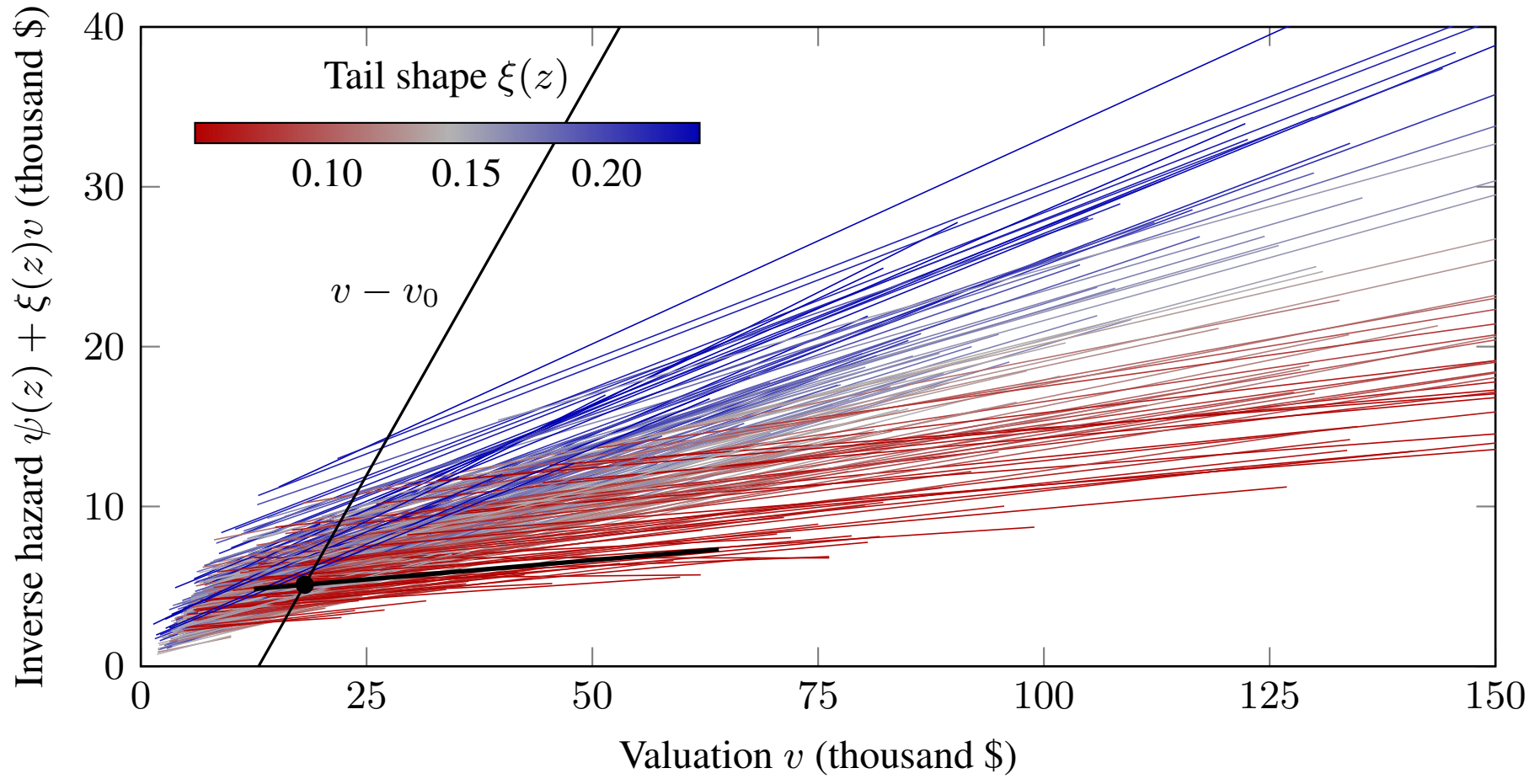
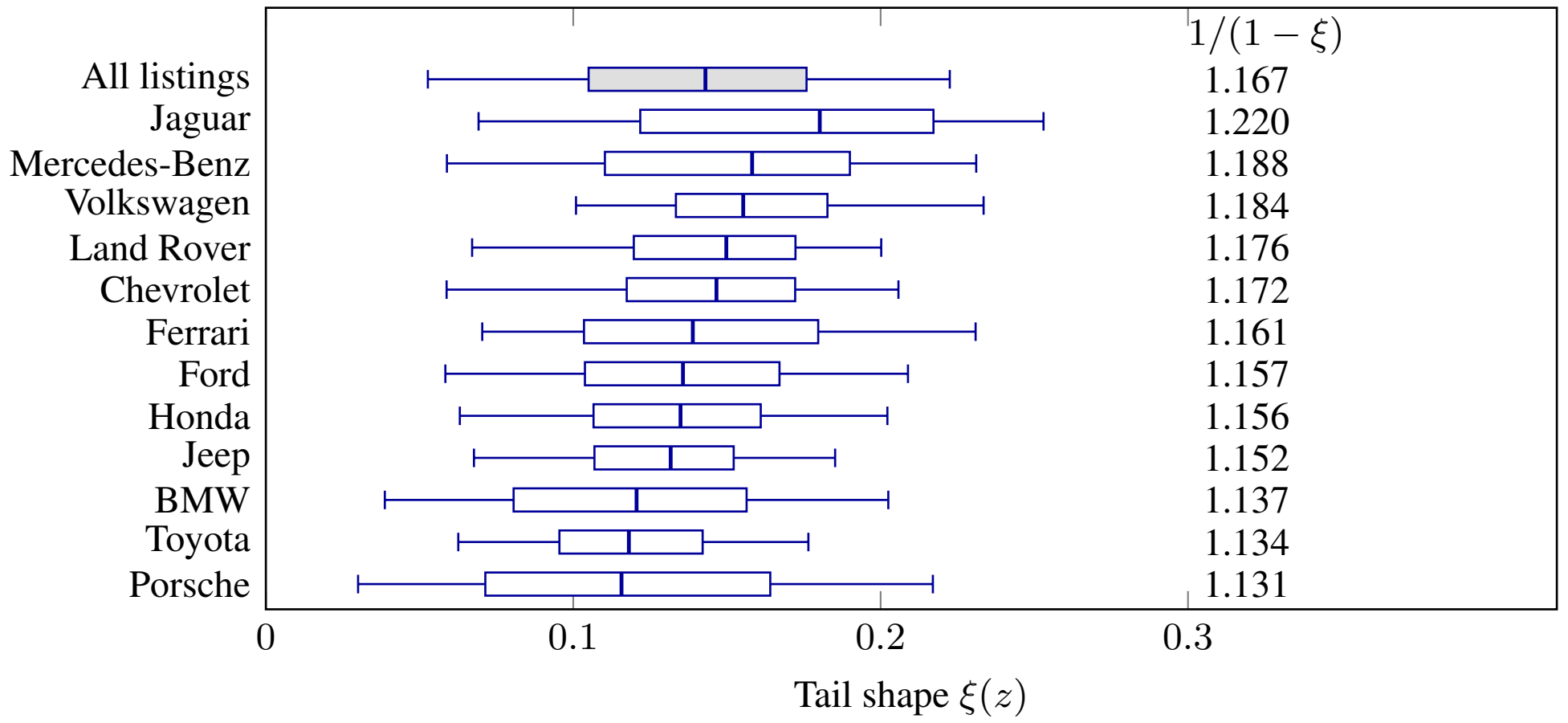

\input{figures/data/tails_styles.tex}
\input{figures/data/tails_panel_b_labels.tex}
\pgfplotsset{
  colormap={divRedWhiteBlueLines}{
    color(0cm)=(red!70!black);
    color(0.5cm)=(white!70!black);
    color(1cm)=(blue!70!black)
  },
  tails axis/.style={
    tick label style={font=\scriptsize},
    label style={font=\footnotesize},
    line width=0.5pt,
    tick style={line width=0.4pt},
    clip=true,
  },
}
\tikzset{tails subcaption/.style={anchor=north, font=\footnotesize, inner sep=0pt, yshift=-3pt}}
\ifdefined\tailsaxiswidth\else\newlength{\tailsaxiswidth}\fi
\begin{center}
\let\tiny\footnotesize
\let\scriptsize\footnotesize
\let\footnotesize\small
\begin{tikzpicture}
  \begin{axis}[
      tails axis,
      name=panelA,
      width=0.74\textwidth, height=0.40\textwidth,
      xlabel={Valuation $v$ (thousand \$)},
      ylabel={Inverse hazard $\psi(z)+\xi(z)v$ (thousand \$)},
      xmin=0, xmax=150, ymin=0, ymax=40,
      xtick={0,25,...,150},
      colormap name=divRedWhiteBlueLines,
      tails panel a meta,
      colorbar horizontal,
      colorbar style={
        at={(0.04,0.85)}, anchor=north west,
        width=0.24\textwidth, height=0.16cm,
        xtick={0.05,0.10,0.15,0.20}, xticklabels={0.05,0.10,0.15,0.20},
        xticklabel style={font=\tiny}, xtick style={draw=none},
        title={Tail shape $\xi(z)$}, title style={font=\tiny, yshift=-4pt},
      },
    ]
    \input{figures/data/tails_panel_a_lines.tex}
    \input{figures/data/tails_featured_overlay.tex}
    \addplot[draw=none, forget plot] coordinates {(0,0)};
  \end{axis}
  \coordinate (osA) at (panelA.outer south);
  \node[tails subcaption] (capA) at (panelA.south |- osA) {(a) Fitted inverse hazards over the identified region};
  \path let \p1=($(panelA.south east)-(panelA.south west)$) in
    \pgfextra{\global\tailsaxiswidth=\x1};
  \begin{axis}[
      tails axis,
      name=panelB,
      at={(panelA.south west |- capA.south)}, anchor=north west, yshift=-14pt,
      scale only axis,
      width=\tailsaxiswidth, height=0.30\textwidth,
      xlabel={Tail shape $\xi(z)$},
      xmin=0, xmax=0.42, ymin=0.4, ymax=14.7,
      xtick={0,0.1,0.2,0.3},
      tails panel b ticks,
      ytick style={draw=none},
      yticklabel style={font=\scriptsize},
      boxplot/draw direction=x,
      boxplot/box extend=0.55,
      boxplot/every median/.style={line width=0.9pt},
    ]
    \node[anchor=west, font=\scriptsize] at (axis cs:0.295,14.05) {$1/(1-\xi)$};
    \input{figures/data/tails_panel_b_boxes.tex}
  \end{axis}
  \coordinate (osB) at (panelB.outer south);
  \node[tails subcaption] at (panelB.south |- osB) {(b) Tail shape by make};
\end{tikzpicture}
\end{center}
  \caption{Estimated Valuation Tails}
  \label{fig:tails}
  \par\smallskip
  \begin{minipage}{0.92\textwidth}
    \footnotesize \textit{Note.} Panel (a) draws the fitted inverse hazard $\psi(z)+\xi(z)v$ of \ResPanelAN{} randomly drawn inference-half listings over the band between the 5th and 95th percentiles of its neighbors' retained thresholds, so each drawn segment lies in the identified region $[\underline{t}(d),\vhigh)$. Lines are colored by the shape $\xi$ on a diverging scale centered at the median shape \ResXiMedian. The reserve of a listing is the valuation at which its inverse hazard meets the line $v-v_0$. The black line is the fitted inverse hazard of the featured listing of Section \ref{sec:token-attribution}, a \ResFeaturedTitle. At the $c^*$ multiplier, its reserve of \$\ResFeaturedReserve{} is marked at the intersection with $v-v_0$, and the bar through it shows the 95\% sampling interval. Panel (b) draws the distribution of the shape by make for the twelve most frequent makes, with boxes at the quartiles, whiskers at the 5th and 95th percentiles, the pooled distribution at the top, and the median tail markup $1/(1-\xi)$ at the right. The tail markup is the pass-through of the seller value to the optimal reserve, $\partial r^{*}/\partial v_{0}=1/(1-\xi)$, so a markup of 1.167 means a one-dollar increase in the seller's value raises the reserve by \$1.167.
  \end{minipage}
\end{figure}

In Figure \ref{fig:tails}, we visually inspect how the two tail parameters add to a price prediction. Panel (a) draws the fitted inverse hazard of a random sample of listings over each listing's support band, which lies inside its identified region, and the reserve of a listing is the point at which that line meets $v-v_0$. The tail markup $1/(1-\xi)$ runs from \ResMarkupQlo{} to \ResMarkupQhi{} between the 5th and 95th percentiles of listings, and the intercept share $\psi/\widehat{\E}[B_{(2)}\mid\hat{z}]$ from \ResInterceptShareQlo{} to \ResInterceptShareQhi. The intercept share falls with the anchor, with a correlation of \ResCorrInterceptAnchor{} against the log anchor, whereas the shape is unrelated to it, with a correlation of \ResCorrXiAnchor. Expensive vehicles therefore carry proportionally smaller intercepts and no thinner tails. Panel (b) shows that the make explains little of the shape. The median shape ranges from \ResMakeLowestXi{} for \ResMakeLowest{} to \ResMakeHighestXi{} for \ResMakeHighest{} across the twelve most frequent makes, a markup range of \ResMakeLowestMarkup{} to \ResMakeHighestMarkup, and between-make variation accounts for only \ResBetweenMakeShare\% of the shape variance among those makes. The tail markup is the pass-through of the seller value to the reserve, $\partial r^{*}/\partial v_{0}=1/(1-\xi(z))$, so a one-dollar increase in the seller's value raises the optimal reserve by between $\ResMarkupQlo{}$ and $\ResMarkupQhi{}$ across these listings. We interpret this result as the tail being a property of the individual listing that its text carries.

\begin{figure}[!htbp]
  \input{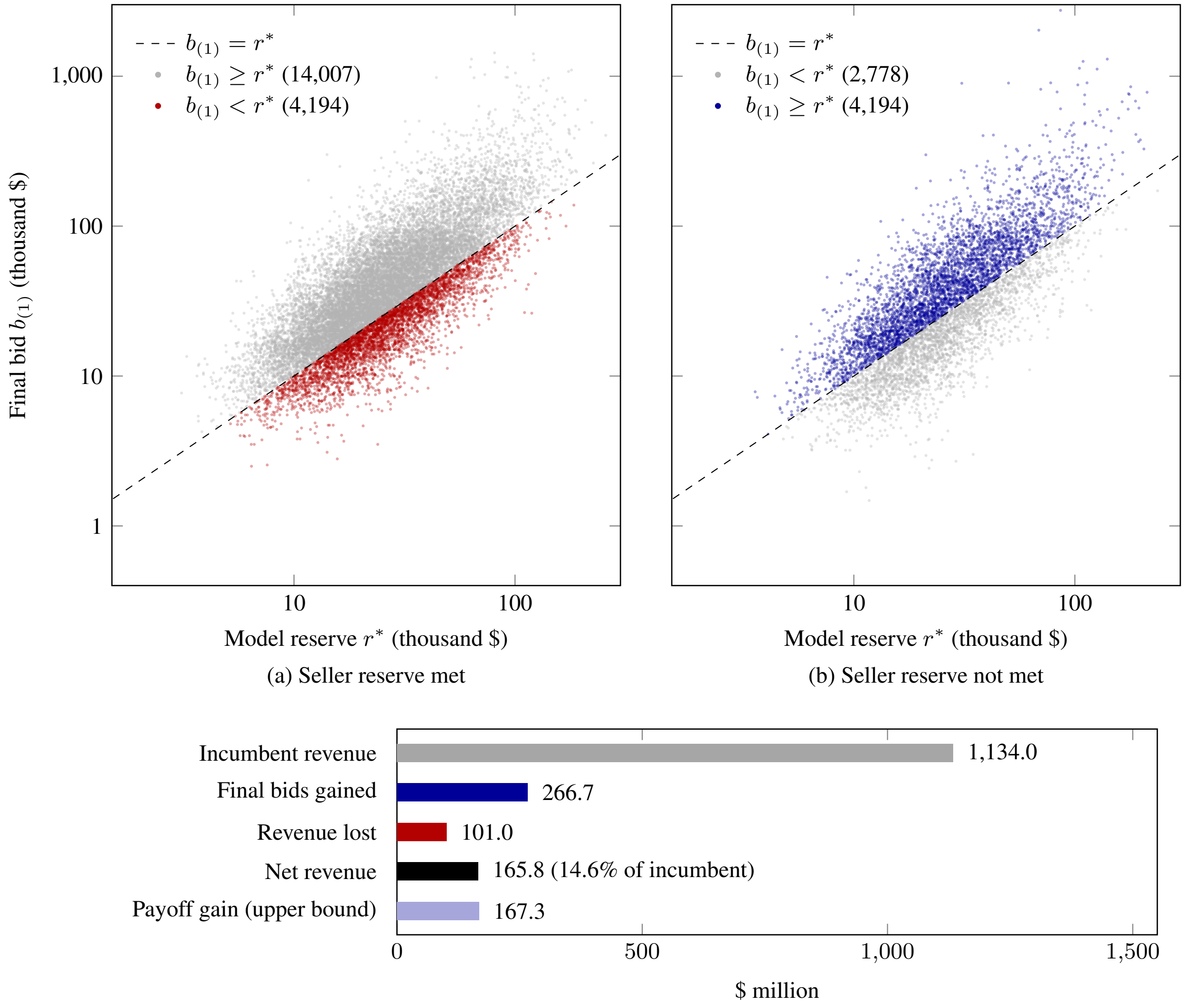}
  \caption{Incumbent Outcomes and Model Reserves}
  \label{fig:reserve-policy}
  \par\smallskip
  \begin{minipage}{0.92\textwidth}
    \footnotesize \textit{Note.} Every reserve listing in $T_{\mathrm{II}}$ is evaluated at the calibrated seller value $v_0=c^\ast\,\widehat{\E}[B_{(2)}\mid\hat{z}]$, $c^\ast=\ResCStar$. Panels (a) and (b) plot the final bid against the model's reserve for the \ResPolicyEligible{} reserve listings, split by whether the seller's reserve was met, with the line $b_{(1)}=r^\ast$ separating auctions whose final bid meets the model's reserve from those whose bid does not. Grey points agree with the model. Colored points disagree with the model. Calibration equates the two disagreement cells among reserve listings. No-reserve outcomes are preserved and omitted from the plots. Panel (c) compares automatic-sale revenue across affected reserve listings. 
  \end{minipage}
\end{figure}

Figure \ref{fig:reserve-policy} compares realized outcomes with the model's reserve price. Panel (a) plots the \ResPolicyEligible{} reserve listings in $T_{\mathrm{II}}$ that sold under the incumbent policy, and panel (b) plots those that did not. Within each panel, the line $b_{(1)}=r^\ast$ separates the auctions on which the two policies agree from those on which they disagree. The disagreements form two groups that mirror each other. In the first group, shown in panel (a), \ResPolNLow{} auctions sold although the final bid fell below $r^\ast$, so the model's higher reserve would have caused them to fail at the auction stage and placed \$\ResPolLowRevenue{} million in realized revenue at risk, depending on the winner's valuation. In the second group, shown in panel (b), \ResPolNHigh{} auctions failed to meet the seller's reserve although the final bid exceeded $r^\ast$, so the model's lower reserve would have converted them into automatic sales worth \$\ResPolHighBidTotal{} million at the observed final bids. The two groups have the same size by construction. Because $c^\ast$ sets the share of final bids that meet $r^\ast$ equal to the incumbent sale rate of \ResMetTargetOne\%, each sale the model's reserve removes in the first group is offset by a sale it adds in the second. No-reserve listings keep their observed outcomes and are omitted from the panels. Panel (c) reports the resulting automatic-sale revenue. The net difference is \$\ResPolNetRevenue{} million, or \ResPolNetIncumbentShare\% of the \$\ResPolIncumbentRevenueBillion{} billion in observed revenue generated by the incumbent policy. Under the full-bargaining continuation of Appendix \ref{app:mechanism}, in which the seller offers $r^\ast$ to the high bidder after the auction, the first group instead yields an expected gain of \$\ResPolTotalGain{} million (Table \ref{tab:bargaining-benchmark}).

\subsection{Sensitivity of Reserve Recommendations to Listing Text}\label{sec:token-attribution}

We examine which parts of a listing description influence its fitted reserve through leave-one-out occlusion at the multiplier $c^*=0.467$. For a text element $j$, we compute $\phi_j=\hat{r}(d)-\hat{r}(d\setminus j)$, where $d\setminus j$ denotes the description with that word or token removed and the remaining text re-encoded through the frozen pipeline. These attributions measure the sensitivity of the fitted recommendation to changes in its input. We stress, however, they do not identify causal effects of product attributes, and interactions among text elements mean that the attributions need not sum to the fitted reserve. To provide an example of this exercise, Figure \ref{fig:token-attribution} shades each of the \ResFeaturedTokens{} content tokens of a randomly selected listing by its dollar-valued $\phi_j$, on a symmetric scale from $-\$\ResFeaturedMaxPhi$ to $\$\ResFeaturedMaxPhi$, around zero change in the fitted reserve of \$\ResFeaturedReserve{}. Among the listing's words, the largest attributions belong to digits. The zeros of the model year ``2001'' that opens the description carry the largest negative attribution, \$\ResFeaturedTopNegPhi{} each, so deleting either raises the reserve by more than a quarter of its level, and the ones of ``911'' in the title carry the largest positive, \$\ResFeaturedTopPosPhi{} each. Deleting either digit of a pair yields the same shortened sequence, so the pair shares one attribution. The attributions sum to \$\ResFeaturedSumPhi, well above the reserve itself, so the deletions interact and the figure is a map of where the reserve is sensitive and not a decomposition of it.

\begin{figure}[!htbp]
  \centering
  \begin{tikzpicture}
    \begin{axis}[
        hide axis, scale only axis, height=0pt, width=0.88\textwidth,
        colorbar horizontal,
        point meta min=-\ResFeaturedMaxPhiRaw,
        point meta max=\ResFeaturedMaxPhiRaw,
        colorbar style={
          width=0.88\textwidth, height=0.4cm,
          xtick={-\ResFeaturedMaxPhiRaw,-\ResFeaturedHalfMaxPhiRaw,0,\ResFeaturedHalfMaxPhiRaw,\ResFeaturedMaxPhiRaw},
          xticklabels={{-\$\ResFeaturedMaxPhi},{-\$\ResFeaturedHalfMaxPhi},{\$0},{\$\ResFeaturedHalfMaxPhi},{\$\ResFeaturedMaxPhi}},
          xlabel={Effect on fitted reserve},
          xticklabel style={text height=1.5ex},
          xtick style={draw=none},
          colormap name=divRedWhiteBlue
        },
      ]
      \addplot [draw=none] coordinates {(-\ResFeaturedMaxPhiRaw,0) (\ResFeaturedMaxPhiRaw,0)};
    \end{axis}
  \end{tikzpicture}

  \raggedright
  {\linespread{0.9}\small
    \input{figures/token-attribution-featured.tex}\par}
  \caption{Token-Level Occlusion Attributions}
  \label{fig:token-attribution}
  \par\smallskip
  \begin{minipage}{0.92\textwidth}
    \footnotesize \textit{Note.} At $c^*=467$, each content token of the featured listing is shaded by its dollar-valued occlusion attribution $\phi_j=\hat r(d)-\hat r(d\setminus j)$. The color bar is centered at zero change and uses a symmetric dollar range. Blue tokens raise the fitted reserve relative to their omission. Red tokens lower it. The end-of-sequence token is omitted.
  \end{minipage}
\end{figure}

Table \ref{tab:token-attribution-corpus} aggregates word-level occlusion over a fixed sample of \ResAttrN{} inference-half listings within the fitted support region. A word is a space-delimited span of tokens deleted whole, case variants are merged, and the panels rank the ten words with the most negative and the most positive mean attributions as a share of the listing's reserve among the \ResAttrWordsQualifying{} words appearing in at least \ResAttrWordsMinListings{} listings. The word ``\ResAttrLowerTopWord'' lowers the reserve by \ResAttrLowerTopPct\% per occurrence across the \ResAttrLowerTopListings{} listings that contain it, and the other nine words of Panel A, condition defects among them, lower it by \ResAttrLowerLastPct\% to \ResAttrLowerSecondPct\%. The words of Panel B raise it by \ResAttrRaiseLastPct\% to \ResAttrRaiseTopPct\%. Template words and the end-of-sequence token were excluded from the analysis in these panels by rule.

\begin{table}[!htbp]
  \centering
  \caption{Corpus Word-Level Occlusion Attributions of the Optimal Reserve}
  \label{tab:token-attribution-corpus}
\begin{tabular}{llrrr}
\toprule
Word & Category & Mean (\$) & Mean (\%) & Listings \\
\midrule
\multicolumn{5}{l}{\emph{Panel A. Words that lower the reserve}} \\
Reserve & reserve status & $-$2,339 & $-$18.3 & 106 \\
1987 & model year & $-$777 & $-$6.9 & 22 \\
blemishes & condition & $-$998 & $-$5.3 & 58 \\
corrosion & condition & $-$779 & $-$4.7 & 45 \\
cracks & condition & $-$768 & $-$4.7 & 27 \\
2001 & model year & $-$728 & $-$4.4 & 28 \\
inline-four & engine & $-$647 & $-$3.9 & 90 \\
dent & condition & $-$593 & $-$3.7 & 24 \\
Toyota & brand & $-$785 & $-$3.6 & 22 \\
Chevrolet & brand & $-$725 & $-$3.4 & 50 \\
\midrule
\multicolumn{5}{l}{\emph{Panel B. Words that raise the reserve}} \\
911 & model & 3,142 & 7.6 & 42 \\
2022 & model year & 2,238 & 7.5 & 182 \\
Sep & other & 1,845 & 7.3 & 49 \\
twin-turbocharged & engine & 2,972 & 6.9 & 33 \\
Nov & other & 1,387 & 6.8 & 49 \\
Jun & other & 1,591 & 5.8 & 43 \\
Land & brand & 1,267 & 5.6 & 29 \\
turbocharged & engine & 1,124 & 4.8 & 30 \\
flat-six & engine & 1,809 & 4.8 & 47 \\
2021 & model year & 1,314 & 4.5 & 183 \\
\bottomrule
\end{tabular}

  \par\smallskip
  \begin{minipage}{0.92\textwidth}
    \footnotesize \textit{Note.} Word-level leave-one-out occlusion at $c^*=0.467$ over \ResAttrN{} inference-half listings selected by fixed rule. Mean (\$) is the mean attribution in dollars under the fitted model and Mean (\%) is the mean attribution as a share of the listing's fitted reserve. Listings is the number of sample listings containing the word. Words qualify with at least \ResAttrWordsMinListings{} listings. Template header words and the end-of-sequence token are excluded.
  \end{minipage}
\end{table}

\subsection{Mileage Gradients in Bids and Recommended Reserves}\label{sec:mileage}\label{sec:counterfactuals}

We examine whether fitted reserves exhibit mileage gradients consistent with those observed in price-setting bids. Mileage provides a useful diagnostic because it is stated in many listings and has a directly interpretable (negative) economic meaning. We use the \ResGradN{} inference-half listings whose descriptions clearly state one odometer value between 5,000 and 200,000 miles and whose model group contains at least \ResGradMinGroup{} such listings. The analysis measures cross-sectional associations conditional on model group and model year, rather than causal effects of mileage on valuations. 

\begin{figure}[!htbp]
\input{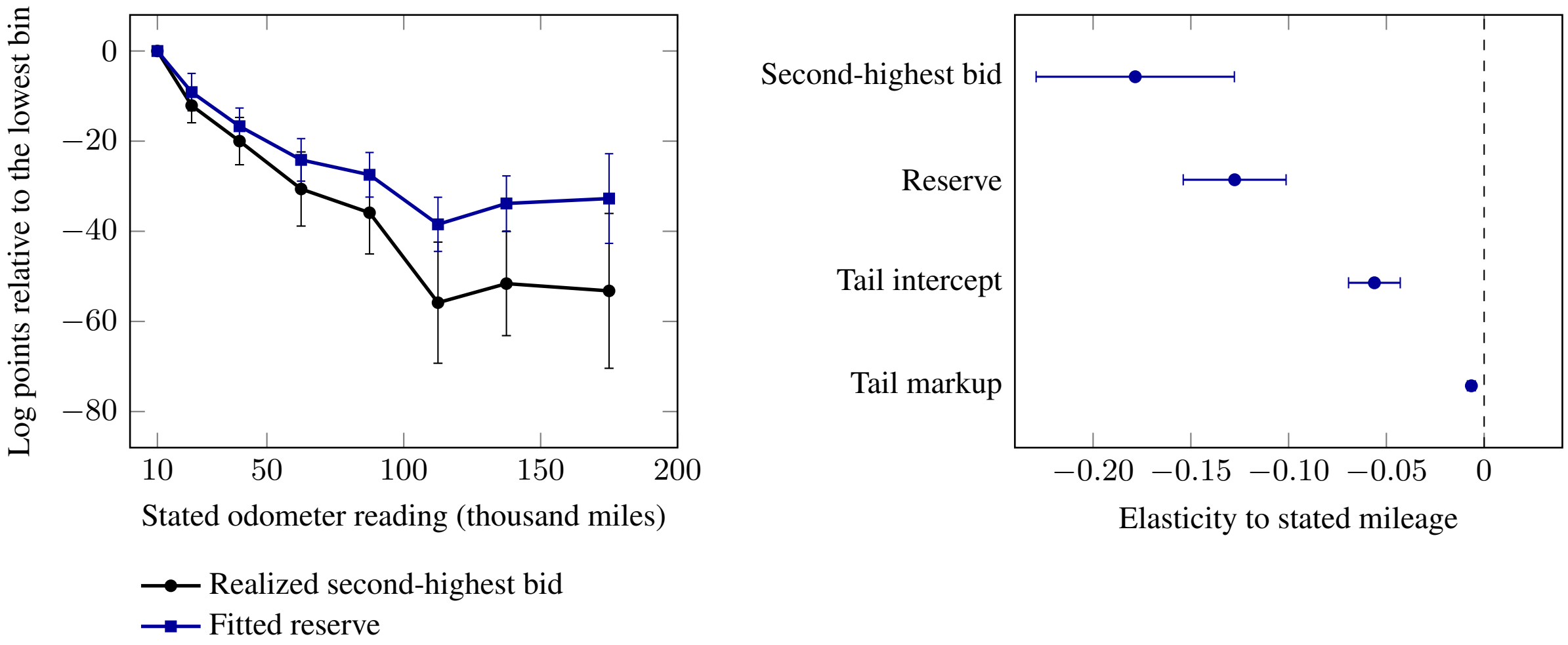}
\pgfplotsset{
  mileage axis/.style={
    tick label style={font=\scriptsize},
    label style={font=\footnotesize},
    line width=0.5pt,
    tick style={line width=0.4pt},
    legend style={font=\scriptsize, draw=none, fill=none},
    legend cell align=left,
  },
}
\tikzset{mileage subcaption/.style={anchor=north, font=\footnotesize, inner sep=0pt}}
\begin{center}
\let\tiny\footnotesize
\let\scriptsize\footnotesize
\let\footnotesize\small
\begin{tikzpicture}
  \begin{groupplot}[
      mileage axis,
      group style={group size=2 by 1, horizontal sep=3.4cm},
      width=0.43\textwidth, height=0.36\textwidth,
    ]
    \nextgroupplot[
      xlabel={Stated odometer reading (thousand miles)},
      ylabel={Log points relative to the lowest bin},
      xmin=0, xmax=200, xtick={10,50,100,150,200},
      ymin=-88, ymax=8,
      legend to name=mileage profile legend,
      legend columns=1,
    ]
    \addplot[black, line width=1pt, mark=*, mark size=1.3pt, error bars/.cd, y dir=both, y explicit]
      table[x=mid_k, y=pct, y error=err_pct, col sep=comma] {figures/data/mileage_profile_b2.csv};
    \addplot[blue!60!black, line width=1pt, mark=square*, mark size=1.2pt, error bars/.cd, y dir=both, y explicit]
      table[x=mid_k, y=pct, y error=err_pct, col sep=comma] {figures/data/mileage_profile_reserve1.csv};
    \legend{{Realized second-highest bid},{Fitted reserve}}
    \nextgroupplot[
      xlabel={Elasticity to stated mileage},
      mileage elasticity ticks,
      ymin=0.4, ymax=4.6,
      ytick style={draw=none},
      yticklabel style={font=\scriptsize},
      xmin=-0.24, xmax=0.04,
      xtick={-0.2,-0.15,-0.1,-0.05,0},
      xticklabels={$-0.20$,$-0.15$,$-0.10$,$-0.05$,$0$},
    ]
    \addplot[black, dashed, line width=0.4pt, forget plot] coordinates {(0,0.4) (0,4.6)};
    \addplot[only marks, mark=*, mark size=1.6pt, blue!60!black, error bars/.cd, x dir=both, x explicit, error bar style={line width=0.6pt}]
      table[x=beta, y=position, x error=halfwidth, col sep=comma] {figures/data/mileage_elasticities.csv};
  \end{groupplot}
  \node[anchor=north, inner sep=0pt] (mileageLegend) at
    ([yshift=-4pt]group c1r1.outer south) {\pgfplotslegendfromname{mileage profile legend}};
  \coordinate (mileageSubcaptionY) at ([yshift=-3pt]mileageLegend.south);
  \node[mileage subcaption] at (group c1r1 |- mileageSubcaptionY)
    {(a) Mileage-bin profiles};
  \node[mileage subcaption] at (group c2r1 |- mileageSubcaptionY)
    {(b) Mileage elasticities};
\end{tikzpicture}
\end{center}
  \caption{Mileage Gradients Within Model Group and Model Year}
  \label{fig:mileage}
  \par\smallskip
  \begin{minipage}{0.92\textwidth}
    \footnotesize \textit{Note.} \ResGradN{} inference-half listings with one stated odometer value between 5,000 and 200,000 miles in a model group with at least \ResGradMinGroup{} such listings. Panel (a) plots the two-way fixed-effects profile of the log realized second-highest bid and of the log fitted reserve by stated-mileage bin, in log points relative to the 5,000 to 15,000-mile bin, with 95\% intervals. Panel (b) plots the elasticity of each log outcome with respect to log mileage within model group and model year, with its 95\% interval. Standard errors are clustered by model group.
  \end{minipage}
\end{figure}

Panel (a) of Figure \ref{fig:mileage} reports the estimates from separate regressions for each outcome $s$ using indicators for eight mileage bins,
\begin{equation}
  \log Y_{is}
  = \alpha_{g(i),s}+\gamma_{t(i),s}
    +\sum_{k=1}^{7}\delta_{ks}\one\{m_i\in I_k\}
    +\varepsilon_{is},
  \label{eq:mileage-profile}
\end{equation}
where $Y_{is}$ is outcome $s$ for listing $i$, $m_i$ its stated mileage, $g(i)$ its model group, and $t(i)$ its model year. We let the model-group and model-year fixed effects enter separately. The omitted bin $I_0$ covers 5,000 to 15,000 miles, and $I_1,\ldots,I_7$ are the remaining bins, with boundaries at 30,000, 50,000, 75,000, 100,000, 125,000, 150,000, and 200,000 miles. We estimate via ordinary least squares with standard errors clustered by model group and fixed effects re-estimated for each outcome and specification. Panel (a) plots $100\hat\delta_{ks}$ for the second-highest bid and the reserve price with 95\% intervals and the omitted bin set to zero. Both bid and reserve profiles decline through 125,000 miles and flatten where the sample thins. The largest step in both is the crossing of 100,000 miles, \ResGradStepHundredBtwo{} log points for realized prices and \ResGradStepHundredReserve{} for the reserve, which is the threshold pattern that \citet{Lacetera2012Heuristic} document in wholesale auctions.

Panel (b) summarizes the mileage gradient with a separate log--log regression for each outcome,
\begin{equation}
  \log Y_{is}
  = \alpha_{g(i),s}+\gamma_{t(i),s}
    +\beta_s\log m_i+\varepsilon_{is},
  \label{eq:mileage-elasticity}
\end{equation}
where $\beta_s$ measures the mileage elasticity of outcome $s$ after accounting for model group and model year. The outcomes are the second-highest bid, fitted reserve, tail intercept $\psi$, and tail markup $1/(1-\xi)$. The elasticity of the realized second-highest bid is \ResGradBtwo{} (standard error \ResGradBtwoSE), and the elasticity of the fitted reserve is \ResGradReserve{}. Thus, the fitted reserve falls with stated mileage at \ResGradShare\% of the rate realized prices do. Between 25,000 and 150,000 miles the implied decline is \ResGradDeclineReserve\% for the reserve and \ResGradDeclineBtwo\% for realized prices.

We show the decline runs through the level of the valuation distribution and not through its tail. The tail intercept $\psi$ has an elasticity of \ResGradPsi{} and the tail markup $1/(1-\xi)$ an elasticity of \ResGradMarkup, so a high-mileage listing has a lower seller value and a lower inverse hazard at every valuation, but its tail is no thinner. We note that the association is cross-sectional within model group and year and is not a causal effect of mileage on valuations.

\section{Conclusion}\label{sec:conclusion}

This paper develops a structural route from an auction description to a listing-specific reserve price. The key identification result shows that consecutive upper order statistics recover the threshold-normalized tail of the valuation distribution without observing or modeling the number of potential bidders. Although the scale of the survival function, the exceedance probability at the lower endpoint of the identified region, remains unidentified, it cancels from the inverse hazard that determines the Riley--Myerson reserve. A generalized Pareto approximation then converts that identified object into a closed-form reserve rule whose parameters vary with a low-dimensional representation of the listing text. The result is a policy tailored to the individual object rather than to a manually defined vehicle category.

The estimation design separates representation learning from structural inference. We fine-tune the text embedding model against the auction likelihood on one half of the sample, freeze the resulting parameters of the fine-tuned language model, and estimate the tail model on the other half. Conditional on the frozen fine-tuned language model, the second stage maps the representation to the parameters of a generalized Pareto tail distribution. After estimation, a new listing requires a forward pass through the language model and fitted structural head to obtain these parameters. The reserve then follows in closed form given the seller's value, without requiring bids for the new listing. A platform can also present the resulting reserve schedule over a range of seller values, making the dependence of the recommendation on this input explicit.

Our main finding addresses where the model would place reserves relative to the auction house's incumbent policy. Preserving no-reserve outcomes and holding the sale rate among reserve listings fixed, lowering reserves in auctions where the observed final bid exceeded the model's reserve would have converted \$\ResPolHighBidTotal{} million in final bids into automatic sales under reserve-invariant bidding. Applying the model's higher reserve elsewhere would have placed \$\ResPolLowRevenue{} million in realized automatic-sale revenue at risk. The resulting net difference is \$\ResPolNetRevenue{} million, or \ResPolNetIncumbentShare\% of incumbent automatic-sale revenue in the inference sample.

The remaining results show how information in the descriptions enters the fitted policy. Condition, engine, and model terms appear among the words with the largest reserve attributions. Within model group and model year, the reserve declines with stated mileage at approximately \ResGradShare\% of the rate observed in the price-setting bids. The mileage gradient operates primarily through the level of valuations rather than through tail thickness. These findings show that the model responds to economically meaningful content that would be costly to encode through a hand-built feature taxonomy.

Several restrictions define the scope of these findings. First, seller value is a policy input, and the headline multiplier is calibrated to match the incumbent sale rate among reserve listings. Both the reserve level and its reported interval are conditional on that choice. Second, the count-free argument requires symmetric independent private values and a valuation distribution that does not vary with market size after conditioning on the description index. Residual common information can correlate bidders' values, while unobserved listing quality can jointly shift valuations and participation, and either violation can prevent bidder count from canceling out of the order-statistic density and bias the estimated reserves. Third, the empirical likelihood uses timing and threshold-crossing rules to select terminal bids that are most likely to reveal valuations. If these rules select auctions or bidder ranks based on unobserved valuations or bidding behavior even after conditioning on the retained threshold and the description's embedding, the estimated valuation tail and resulting reserves may be biased. And fourth, the base embedding model's pretraining corpus is not fully observable and may include platform listings or bid histories from the study period, so the sample split cannot rule out prior exposure to inference-half observations. The representation is therefore fixed during the second stage but not demonstrably independent of the inference sample, and the individual-listing intervals condition on this fixed representation. These qualifications are central to implementation. 

The broader implication of this work is that unstructured product information can support economically disciplined, item-level decisions without first being reduced to a hand-built feature taxonomy. For operations management, this points toward a broader use of natural language processing models as inputs to structural decision models rather than as ends in themselves. The same approach can inform pricing, matching, procurement, assortment, and other operational decisions in markets where products are heterogeneous and much of their relevant information is recorded in text or images. Across these settings, the central research challenge is to determine which features of latent demand are identified by observed behavior and how uncertainty in a learned representation propagates into the resulting policy. Progress will require designs that align representation learning with the economic objective while preserving a clear link between the data, the behavioral model, and the decision rule. This paper offers one example of that agenda by showing how auction theory, bid histories, and product descriptions can be combined to produce a transparent and listing-specific policy.

\clearpage
\appendix
\numberwithin{equation}{section}
\section{Proofs}\label{app:proofs}

\begin{proof}[Proof of Lemma \ref{lem:song}]
Fix $d$ and $n\geq k$ with positive conditional probability given $d$, and let $v_{(k)}$ be in the support of $V_{(k)}$ given $N=n$ and $d$. All conditioning on the continuously distributed event $V_{(k)}=v_{(k)}$ below is understood through its regular conditional density. Under the valuation model of Section \ref{sec:distribution}, the $n$ valuations are independent draws from $F(\cdot\mid d)$ given the count and description. By the Markov property of order statistics, the $k-1$ valuations above $v_{(k)}$, namely $V_{(1)},\ldots,V_{(k-1)}$, are the descending arrangement of $k-1$ independent draws from $F(\cdot\mid d)$ truncated below at $v_{(k)}$ \citep{Arnold2008First}. Their ordered joint density is
\begin{equation*}
  \frac{(k-1)!\prod_{j=1}^{k-1}f(v_{(j)}\mid d)}{S(v_{(k)}\mid d)^{k-1}}
  \one\{v_{(1)}>\cdots>v_{(k-1)}>v_{(k)}\}.
\end{equation*}
The highest valuation is not observed. Integrating $v_{(1)}$ from $v_{(2)}$ to $\vhigh$ replaces $f(v_{(1)}\mid d)$ with $S(v_{(2)}\mid d)$ and gives equation \eqref{eq:song}. The expression does not involve $n$, so averaging over the conditional law of the count given $N\geq k$ leaves it unchanged.

For the uniqueness claim, hold $(d,v_{(k)})$ fixed and define the threshold-normalized survival function and its density
\begin{equation*}
  G_{d,v_{(k)}}(v)=\frac{S(v\mid d)}{S(v_{(k)}\mid d)},
  \qquad
  g_{d,v_{(k)}}(v)=-G_{d,v_{(k)}}'(v),
  \qquad v\geq v_{(k)}.
\end{equation*}
The numerator of equation \eqref{eq:song} carries $k-1$ factors of the parent law, namely $S(v_{(2)}\mid d)$ together with the $k-2$ densities $f(v_{(2)}\mid d),\ldots,f(v_{(k-1)}\mid d)$, and the denominator carries $k-1$ factors of $S(v_{(k)}\mid d)$. Assigning one denominator factor to each numerator factor leaves $G_{d,v_{(k)}}(v_{(2)})$ and $g_{d,v_{(k)}}(v_{(j)})$ for $j=2,\ldots,k-1$, since $S(v_{(k)}\mid d)$ does not depend on $v$, so
\begin{equation}
  f_{2,\dots,k}(v_{(2)},\ldots,v_{(k-1)}\mid v_{(k)},d)
  =(k-1)!\,G_{d,v_{(k)}}(v_{(2)})\prod_{j=2}^{k-1}g_{d,v_{(k)}}(v_{(j)})
  \one\{v_{(2)}>\cdots>v_{(k-1)}>v_{(k)}\}.
  \label{eq:songnormalized}
\end{equation}

Equation \eqref{eq:songnormalized} first shows that $G_{d,v_{(k)}}$ determines the conditional density. Conversely, integrating it over $v_{(2)}>\cdots>v_{(k-1)}>v$ gives equation \eqref{eq:conditionalsurvivalratio}, so
\begin{equation*}
  G_{d,v_{(k)}}(v)=\Pr\!\left(V_{(k-1)}>v\mid V_{(k)}=v_{(k)},d\right)^{1/(k-1)},
  \qquad v\geq v_{(k)}.
\end{equation*}
The right-hand side is a functional of the conditional law, and the $(k-1)$-th root is single valued on $[0,1]$. Thus the conditional law and $G_{d,v_{(k)}}$ determine one another on $[v_{(k)},\vhigh)$, which is the uniqueness claim.
\end{proof}

\begin{proof}[Proof of Proposition \ref{prop:policyid}]
\emph{Part (i).} Fix $d$. By ancillary retention, the retained conditional densities equal \eqref{eq:song}. The experiment uses only these densities at thresholds $v_{(k)}\in[\underline{t}(d),\vhigh)$ and leaves the marginal law of thresholds and retained depths unspecified. Suppose $\tilde S=\lambda S(\cdot\mid d)$ on $[\underline{t}(d),\vhigh)$ for some $\lambda>0$. Then $\tilde f=\lambda f(\cdot\mid d)$ on the same interval, so $\lambda$ cancels from equation \eqref{eq:song} and the two parents imply the same conditional densities. Conversely, equality of those densities and integration over $v_{(2)}>\cdots>v_{(k-1)}>v$ give
\[
  \frac{\tilde S(v)}{\tilde S(v_{(k)})}
  =\frac{S(v\mid d)}{S(v_{(k)}\mid d)}
  \qquad v\geq v_{(k)}
\]
at every threshold $v_{(k)}\in[\underline{t}(d),\vhigh)$. The ratio $\tilde S/S$ is therefore constant above each available threshold. These intervals overlap and their lower endpoints approach $\underline{t}(d)$, so the constants agree. Continuity extends proportionality to $\underline{t}(d)$. If $\underline{t}(d)=\vlow$, then $\tilde S(\vlow)=S(\vlow\mid d)=1$ forces $\lambda=1$. If $\underline{t}(d)>\vlow$, any $\lambda\in(0,1/S(\underline{t}(d)\mid d))$ with $\lambda\neq1$ allows a continuous extension below $\underline{t}(d)$ with a strictly positive density, which produces a distinct conditionally equivalent parent.

\emph{Part (ii).} Part (i) implies that every conditionally equivalent parent has survival $\lambda S$ and density $\lambda f$ above $\underline{t}(d)$. The scale cancels from $S/f$, so the inverse hazard is common to the entire equivalence class. The strictly positive interior density makes this ratio well defined at every interior point of the identified region.

\emph{Part (iii).} Fix a conditionally equivalent parent $\tilde F$ and $n\geq1$. In the English-auction equilibrium with a publicly announced, committed reserve \citep{vickrey1961counterspeculation,wolfstetter1996auctions}, a seller with value $v_0$ facing $n$ bidders whose valuations are drawn from $\tilde F$ and setting reserve $r$ has expected payoff
\begin{equation}
  \begin{aligned}
    \Pa_n(r,\tilde F)
    ={}&v_0\tilde F(r)^{n}
    +nr\tilde S(r)\tilde F(r)^{n-1}\\
    &+n(n-1)\int_r^{\vhigh}
    v\tilde F(v)^{n-2}\tilde S(v)\tilde f(v)\,dv.
  \end{aligned}
  \label{eq:benchmarkrevenue}
\end{equation}
The first term is the seller's value when every valuation falls below the reserve, the second term is reserve revenue when exactly one valuation exceeds it, and the third term is the expected second-highest valuation when at least two exceed it. For $n=1$ the third term is absent. The finiteness hypothesis $\int_{\underline{t}(d)}^{\vhigh}S(v\mid d)\,dv<\infty$ guarantees that the integral converges for every $r>\vlow$, because the integrand is bounded on $[r,\underline{t}(d)]$. Differentiating equation \eqref{eq:benchmarkrevenue} gives
\begin{equation}
  \frac{\partial\Pa_n(r,\tilde F)}{\partial r}
  =n\tilde F(r)^{n-1}\bigl[\tilde S(r)-(r-v_0)\tilde f(r)\bigr].
  \label{eq:benchmarkderivative}
\end{equation}
The bracket does not depend on $n$, and the leading factor $n\tilde F(r)^{n-1}$ is strictly positive for every $n\geq1$ and $r>\vlow$. Write $\phi(r)=r-\tilde S(r)/\tilde f(r)$ for the virtual valuation, so that the bracket equals $\tilde f(r)\bigl[v_0-\phi(r)\bigr]$. For $r<v_0$ the bracket is strictly positive under every parent, so no maximizer lies below $v_0$. Above $\underline{t}(d)$, part (i) gives $\tilde S=\lambda S(\cdot\mid d)$ and $\tilde f=\lambda f(\cdot\mid d)$ with $\lambda>0$, so $\phi$ coincides with the identified virtual valuation $r-S(r\mid d)/f(r\mid d)$, and by continuity $\phi(\underline{t}(d))$ is identified. If $v_0<\underline{t}(d)$, monotonicity of $\phi$ on the entire support gives $\phi(r)\leq\phi(\underline{t}(d))<v_0$ for every $r\in[v_0,\underline{t}(d)]$, so the bracket is strictly positive on that interval under every parent in the class. The sign of the bracket is therefore identified at every $r\geq v_0$, and it depends neither on which member of the equivalence class generates the data nor on the law of $N$. Let $M$ be the set of $r\in(\max\{v_0,\underline{t}(d)\},\vhigh)$ at which $\phi(r)=v_0$. Monotonicity of $\phi$ makes $M$ an interval, possibly a single point, with the bracket strictly positive on $[v_0,\inf M)$ and strictly negative above $\sup M$. The bracket is positive just above $v_0$, since $\tilde S(v_0)>0$ and the density is continuous. If $M$ were empty, continuity would make the bracket positive throughout $(v_0,\vhigh)$, so $(r-v_0)\tilde S(r)$ would be strictly increasing. This contradicts its limit of zero at a finite upper endpoint. At an infinite upper endpoint, it implies $\tilde S(r)\geq C/(r-v_0)$ for all sufficiently large $r$ and some $C>0$, contradicting the finite tail integral. Thus $M$ is nonempty. For every $n\geq1$, $\Pa_n(\cdot,\tilde F)$ is then strictly increasing on $[v_0,\inf M]$, constant on $M$, and strictly decreasing on $[\sup M,\vhigh)$, so $M$ is its maximizer set.

The seller chooses $r$ before $N$ is realized. For every $r\in[v_0,\vhigh)$ and every $n\geq1$ write
\begin{equation*}
  \Delta_n(r)=\Pa_n(r,\tilde F)-\Pa_n(v_0,\tilde F)
  =\int_{v_0}^{r}n\tilde F(s)^{n-1}\bigl[\tilde S(s)-(s-v_0)\tilde f(s)\bigr]\,ds,
\end{equation*}
and set $\Delta_0\equiv0$, because a seller facing no bidders retains $v_0$ at every reserve. Since $nx^{n-1}(1-x)\leq1$ on $[0,1]$ and $\bigl|\int_{v_0}^{r}(s-v_0)\,n\tilde F(s)^{n-1}\tilde f(s)\,ds\bigr|\leq|r-v_0|$, every $\Delta_n$ satisfies $|\Delta_n(r)|\leq2|r-v_0|$. The expected gain over the reserve $v_0$,
\begin{equation*}
  \E_N\!\left[\Delta_N(r)\right]
  =\sum_{n\geq0}\Pr(N=n\mid d)\,\Delta_n(r),
\end{equation*}
therefore converges absolutely for every law of $N$ given $d$ and requires no moment restriction on the count, and it differs from $\E_N[\Pa_N(r,\tilde F)]$ by a constant whenever the latter is finite. Each $\Delta_n$ with $n\geq1$ is strictly increasing on $[v_0,\inf M]$, constant on $M$, and strictly decreasing on $[\sup M,\vhigh)$, so the mixture is nondecreasing, constant, and nonincreasing on the same three sets, and strictly so on the first and the third because $\Pr(N\geq1\mid d)>0$. The maximizer set of the expected gain is therefore $M$ for every conditionally equivalent parent and every law of $N$ given $d$, which is the count-free claim. When $M$ is a single interior point, the bracket vanishes there, which is equation \eqref{eq:myerson1} with $F(\cdot\mid d)$ in place of $F$.
\end{proof}

\subsection{Standard Errors Conditional on the Index}
\label{sec:conditional-target}

Let $\mathcal{F}_{\mathrm{I}}$ be the information generated by everything Stage I uses, including the pre-trained embedding model, the auctions in $T_{\mathrm{I}}$, and the initialization, data order, and hyperparameter search of fine-tuning. All expectations, variances, and limits in this subsection are conditional on $\mathcal{F}_{\mathrm{I}}$, under which the frozen index $\hat{z}$ is a fixed function of text and the auctions in $T_{\mathrm{II}}$ are independent draws from the population. The population coordinate means and standard deviations of $\hat{z}$, together with the head coefficients that maximize the expected conditional log likelihood, define $\vartheta^{\dagger}=(\mu^{\dagger},\tau^{\dagger},\gamma^{\dagger})$, and the reserve target is
\begin{equation}
  r^{\dagger}(v_0\mid d)
  =\frac{v_0+\psi_{\vartheta^{\dagger}}(\hat{z}(d))}{1-\xi_{\vartheta^{\dagger}}(\hat{z}(d))}.
  \label{eq:target}
\end{equation}
The estimate $\hat r(v_0\mid d)$ replaces $\vartheta^{\dagger}$ with $\hat\vartheta$. Each frozen representation defines its own target, so \eqref{eq:target} requires no assumption that $\hat{z}$ approaches a population limit.

Let $n=|T_{\mathrm{II}}|$, let $X$ contain an auction's description, retained depth, and bids, and write $Z=\hat z(D)$. Let $m(X,\vartheta)$ be the negative log likelihood contribution in \eqref{eq:gpdloglik} with tail parameters $h_{\vartheta}(Z)$, set to $+\infty$ outside the likelihood domain. The estimate $\hat\vartheta$ solves $\bar g_n(\hat\vartheta)=0$, where $\bar g_n(\vartheta)=n^{-1}\sum_{t\in T_{\mathrm{II}}}g(X_t,\vartheta)$ and
\begin{equation*}
  g(X,\vartheta)=
  \begin{pmatrix}
    Z-\mu\\
    (Z-\mu)^{\odot2}-\tau^{\odot2}\\
    \nabla_{\gamma}m(X,\vartheta)
  \end{pmatrix},
\end{equation*}
with $\odot2$ denoting coordinate-wise squaring. The first $2p$ coordinates define the standardization constants, and the remaining $2(p+1)$ coordinates are the first-order conditions of \eqref{eq:stage2}. Stacking the standardization moments with the head score propagates the sampling variation in $(\hat\mu,\hat\tau)$ into the head and the reserve. Let $J$ be the expected Jacobian of $g$ at $\vartheta^{\dagger}$ and $\Omega=\operatorname{Var}(g(X,\vartheta^{\dagger})\mid\mathcal F_{\mathrm I})$. Under the extremum and estimating-equation conditions of \citet{Newey1994Large}, applied to the conditional population law, $\hat\vartheta$ is consistent for $\vartheta^{\dagger}$ and
\begin{equation*}
  \sqrt n\,\bigl(\hat\vartheta-\vartheta^{\dagger}\bigr)
  \;\xrightarrow{d}\;
  \mathcal N\!\left(0,\;J^{-1}\Omega\bigl(J^{-1}\bigr)^{\prime}\right).
\end{equation*}
Correct specification is not assumed, so the information matrix equality need not hold and the variance retains its sandwich form \citep{Huber1967Behavior,White1982Maximum}. In that case, $\gamma^{\dagger}$ minimizes the expected Kullback--Leibler divergence between the modeled and actual conditional laws. The conditions require a unique interior population minimizer, moment and continuity conditions that accommodate a likelihood that is infinite at parameters whose support excludes an observation, and a finite nonsingular $J$. The generalized Pareto support moves with the parameters, so smooth links alone do not imply these conditions. The shape link bounds $\xi$ below by $-0.45$, which places every admissible tail in the region $\xi>-1/2$ where \citet{Smith1985Maximum} establishes regular root-$n$ asymptotics for the generalized Pareto likelihood, and we impose the corresponding conditions on the conditional model. The empirical Jacobian and domain checks in Appendix \ref{app:computation} are diagnostics for these conditions and do not establish them.

Standard errors for the reserve follow from the delta method. Let $a$ be the gradient of \eqref{eq:target} with respect to $\vartheta$ at $\vartheta^{\dagger}$, which exists because the standardization constants are positive and the links are smooth and keep $\xi_{\vartheta}$ below one, and suppose $s^2=a^{\prime}J^{-1}\Omega(J^{-1})^{\prime}a>0$. Then $\sqrt n\,(\hat r(v_0\mid d)-r^{\dagger}(v_0\mid d))\xrightarrow{d}\mathcal N(0,s^2)$. The reported standard error is $\widehat{\mathrm{se}}=n^{-1/2}\bigl[\hat a^{\prime}\hat J^{-1}\hat\Omega(\hat J^{-1})^{\prime}\hat a\bigr]^{1/2}$, with $\hat a$, $\hat J$, and $\hat\Omega$ the sample counterparts at $\hat\vartheta$, and the interval $\hat r(v_0\mid d)\pm q_{1-\alpha/2}\widehat{\mathrm{se}}$ contains $r^{\dagger}(v_0\mid d)$ with conditional probability approaching $1-\alpha$. If the conditions hold for almost every realization of $\mathcal F_{\mathrm I}$, dominated convergence of the conditional coverage probability gives unconditional coverage of $1-\alpha$ for the target associated with the realized representation.

The argument follows the logic of inference after sample splitting \citep{Cox1975Note}. A model chosen on one part of the data can be assessed on an independent part as if it had been fixed in advance. Related uses of this device include inference for parameters defined by a selected model without assuming that model is correct \citep{Rinaldo2019Bootstrapping}, honest estimation within a data-adaptive partition \citep{Athey2016Recursive}, and inference for features of machine-learning proxies that need not be consistent \citep{Chernozhukov2025Generic}. Our representation plays the role of the selected model, partition, or proxy.

Under the maintained auction assumptions, correct specification, and Assumption \ref{ass:sufficiency} for $\hat z$, the target \eqref{eq:target} equals the optimal reserve on the domain of \eqref{eq:gpdreserve}. The inference does not require this structural interpretation. If the index discards valuation-relevant information, the best fitting head can remain systematically different from the description-specific valuation law even with an arbitrarily large inference sample. Sample splitting prevents inference-half outcomes from influencing fine-tuning or model selection, but it does not recover information discarded by the index, and the sandwich variance accommodates misspecification around $\vartheta^{\dagger}$ without correcting the difference between $r^{\dagger}$ and the economic optimum. This distinction is related to the difference between predictive fine-tuning and structural estimation studied by \citet{vafa2025estimating}. Stage I uses the same conditional auction likelihood as Stage II, so the representation is trained on the valuation-tail model itself. This alignment does not establish Assumption \ref{ass:sufficiency}. Orthogonal-score methods address sensitivity to nuisance estimation when estimating a common population target \citep{Chernozhukov2018Double,Farrell2021Deep}. Our likelihood score is not constructed to be orthogonal to changes in the representation, and we instead condition on the learned index and conduct inference for its associated population head. Inference for the true optimal reserve without sufficiency and correct specification would require further identification and bias-control arguments.

Conditional independence of the inference half from the index also requires that no inference-half outcome entered the embedding model's pretraining \citep{Ludwig2025Large}. We cannot verify this condition for the base model. Sample splitting protects against reuse of inference-half outcomes during fine-tuning but does not establish their absence from pretraining.

\section{The Platform Mechanism and Commitment Equivalence}\label{app:mechanism}

Proposition \ref{prop:policyid}(iii) prices a publicly announced, committed reserve, whose payoff is \eqref{eq:benchmarkrevenue} in Appendix \ref{app:proofs}, whereas the platform privately compares its reserve with the final standing price $b_{(1)}$ and routes an unmet auction to negotiation. The result below uses the continuous-bid benchmark in which that price equals $V_{(2)}$. In the data, bid increments separate the winning bid from the second-highest terminal bid, which averages 97.5\% of the winning bid (Table \ref{tab:summary-1}). Applying the result empirically therefore treats that gap as negligible. Subject to this approximation, two conditions make the platform rule equivalent to the committed benchmark. We then relax the bargaining condition and bound the resulting reserve.

\begin{assumption}[Reserve-invariant bidding and the negotiation continuation]
  \phantomsection
  \label{ass:mechanism}
  \leavevmode
  \begin{enumerate}
    \item[(a)] Bidding is invariant to the secret reserve, and in the continuous-bid benchmark the final standing price is $b_{(1)}=V_{(2)}$.
    \item[(b)] When $b_{(1)}<r$, the seller makes a take-it-or-leave-it offer at $r$ in the ensuing negotiation, which the high bidder accepts if and only if $V_{(1)}\geq r$. Otherwise the seller retains the object at $v_0$.
  \end{enumerate}
\end{assumption}

Part (a) combines a weak-dominance argument with the continuous-bid approximation stated above. Because the reserve is neither announced nor used to set the price, it does not enter any bidder's payoff from remaining active, so bidding up to one's valuation is weakly dominant, as it is with no reserve at all \citep{vickrey1961counterspeculation,wolfstetter1996auctions}. The valuation law $F(\cdot\mid z)$ is therefore invariant to the reserve policy. Part (b) assigns the seller full bargaining power at the posted reserve in the continuation the platform operates, and it is the restriction on which the next result rests. The seller also pays a listing fee $\kappa$ regardless of the outcome, \$99 in the data.

\begin{proposition}[Commitment equivalence]
  \label{prop:commitment}
  Under the valuation model of Section \ref{sec:distribution}, let Assumption \ref{ass:mechanism} hold and let $\Pa^{\mathrm{BaT}}_n(r,F)$ denote the seller's expected payoff under the platform's secret-reserve rule with $n\geq1$ bidders. Then for every $r$ and every $n\geq1$,
  \begin{equation*}
    \Pa^{\mathrm{BaT}}_n(r,F)=\Pa_n(r,F)-\kappa,
  \end{equation*}
  where $\Pa_n$ is the committed-reserve payoff \eqref{eq:benchmarkrevenue}. The two mechanisms therefore have the same optimal reserves, and Proposition \ref{prop:policyid}(iii) applies to the platform's rule unchanged.
\end{proposition}

\begin{proof}
Fix $n\geq1$ and $r$, and adopt the convention $V_{(2)}=\vlow$ when $n=1$. Under Assumption \ref{ass:mechanism}(a), $b_{(1)}=V_{(2)}$ and the valuation vector is unaffected by $r$, so the following events partition the outcome space identically under both mechanisms. Write $\Pa^{\mathrm{BaT}}_n$ for the seller's payoff under the platform mechanism, gross of the listing fee.

\emph{Case 1, $V_{(2)}\geq r$.} The reserve is met, the platform completes the sale at $b_{(1)}=V_{(2)}$, and the seller receives $V_{(2)}$. Under the committed reserve at least two valuations exceed $r$, so the seller again receives $V_{(2)}$. Integrating over this event gives the third term of \eqref{eq:benchmarkrevenue}, since $n(n-1)F(v)^{n-2}S(v)f(v)$ is the density of $V_{(2)}$. This term is absent when $n=1$.

\emph{Case 2, $V_{(1)}\geq r>V_{(2)}$.} The reserve is not met, and the platform routes the seller and the high bidder to negotiation. By Assumption \ref{ass:mechanism}(b), the seller offers $r$ and the high bidder accepts, so the seller receives $r$. Under the committed reserve exactly one valuation exceeds $r$, and the seller also receives $r$. This event has probability $nS(r)F(r)^{n-1}$ and gives the second term of \eqref{eq:benchmarkrevenue}.

\emph{Case 3, $V_{(1)}<r$.} The offer at $r$ is rejected and the seller retains the object at $v_0$, as under the committed reserve. This event has probability $F(r)^n$ and gives the first term of \eqref{eq:benchmarkrevenue}.

Summing, $\Pa^{\mathrm{BaT}}_n(r,F)$ gross of fees equals $\Pa_n(r,F)$. The listing fee $\kappa$ is charged in every one of the three cases and does not depend on $r$, so it enters the seller's payoff as an additive constant. Hence $\Pa^{\mathrm{BaT}}_n(r,F)=\Pa_n(r,F)-\kappa$ for every $r$ and $n$, the two objectives have identical maximizer sets, and Proposition \ref{prop:policyid}(iii) applies to the platform's rule unchanged.
\end{proof}

The equivalence rests on the intermediate state $V_{(1)}\geq r>V_{(2)}$, where Assumption \ref{ass:mechanism}(b) makes negotiation deliver the same $r$ as commitment. The following extension weakens that condition.

\subsection{Bargaining Power in the Continuation}\label{sec:bargaining}
Replace Assumption \ref{ass:mechanism}(b) by the reduced form in which the seller's payoff in Case 2 above is $v_0+\lambda(r-v_0)$ for some $\lambda\in[0,1]$, so that $\lambda=1$ is Assumption \ref{ass:mechanism}(b) and $\lambda=0$ is a negotiation returning the seller's outside option. Cases 1 and 3 are unchanged, so
\begin{equation*}
  \Pa_n(r,\lambda)
  =v_0F(r)^{n}
  +n\bigl[v_0+\lambda(r-v_0)\bigr]S(r)F(r)^{n-1}
  +n(n-1)\int_r^{\vhigh}vF(v)^{n-2}S(v)f(v)\,dv,
\end{equation*}
suppressing the conditioning on $z$. Differentiating and collecting the terms in $(r-v_0)$ gives
\begin{equation}
  \frac{\partial\Pa_n(r,\lambda)}{\partial r}
  =n\lambda F(r)^{n-1}\bigl[S(r)-(r-v_0)f(r)\bigr]
  -n(n-1)(1-\lambda)S(r)F(r)^{n-2}f(r)(r-v_0),
  \label{eq:lambdafoc}
\end{equation}
which reduces to \eqref{eq:benchmarkderivative} at $\lambda=1$. Write $D(r)=S(r)-(r-v_0)f(r)$ for the bracket in the first term.

\begin{corollary}[Bargaining power and the optimal reserve]
  \label{cor:bargaining}
  Under the valuation model of Section \ref{sec:distribution}, let Assumption \ref{ass:mechanism}(a) hold, fix $n\geq2$, and suppose the virtual valuation $v-S(v)/f(v)$ is strictly increasing, so that $D(r)=0$ has a unique solution $r^{*}_{1}>v_0$, the committed-reserve optimum of \eqref{eq:myerson1}. Under the reduced-form continuation with bargaining weight $\lambda\in[0,1]$,
  \begin{enumerate}
    \item[(i)] at $\lambda=0$ the unique optimal reserve is $v_0$,
    \item[(ii)] for every $\lambda\in[0,1]$ no optimal reserve exceeds $r^{*}_{1}$, and
    \item[(iii)] there is a selection of optimal reserves that is nondecreasing in $\lambda$ and runs from $v_0$ at $\lambda=0$ to $r^{*}_{1}$ at $\lambda=1$.
  \end{enumerate}
\end{corollary}

\begin{proof}
Strict monotonicity of the virtual valuation makes $D$ strictly positive below $r^{*}_{1}$ and strictly negative above it. On $(\vlow,v_0)$ both terms of \eqref{eq:lambdafoc} are nonnegative, the first strictly so when $\lambda>0$ and the second strictly so when $\lambda<1$, and the payoff is constant in $r$ below $\vlow$, so no optimal reserve lies below $v_0$ at any $\lambda$.

\emph{Part (i).} At $\lambda=0$ the derivative equals $-n(n-1)S(r)F(r)^{n-2}f(r)(r-v_0)$, which is strictly negative for $r>v_0$, so the payoff is strictly decreasing above $v_0$ and $v_0$ is its unique maximizer.

\emph{Part (ii).} For $r>r^{*}_{1}$ both terms of \eqref{eq:lambdafoc} are nonpositive, the first strictly so when $\lambda>0$ and the second strictly so when $\lambda<1$, so the payoff is strictly decreasing above $r^{*}_{1}$ and no maximizer exceeds it.

\emph{Part (iii).} By the preceding paragraphs every maximizer lies in $[v_0,r^{*}_{1}]$ at every $\lambda$. On that interval
\begin{equation*}
  \frac{\partial^{2}\Pa_n(r,\lambda)}{\partial r\,\partial\lambda}
  =nF(r)^{n-1}D(r)+n(n-1)S(r)F(r)^{n-2}f(r)(r-v_0)\;\geq\;0,
\end{equation*}
since $D\geq0$ there, so the payoff has increasing differences in $(r,\lambda)$ on $[v_0,r^{*}_{1}]\times[0,1]$ and the monotonicity theorem of \citet{topkis1978minimizing} delivers a nondecreasing selection of maximizers. Its endpoints are $v_0$ at $\lambda=0$ by part (i) and $r^{*}_{1}$ at $\lambda=1$, where \eqref{eq:lambdafoc} reduces to \eqref{eq:benchmarkderivative} and takes the sign of $D$.
\end{proof}

With one bidder the payoff at $\lambda=0$ is constant in $r$, so $v_0$ remains optimal but not uniquely so. For $0<\lambda<1$, the factor $n(n-1)$ in the second term of \eqref{eq:lambdafoc} does not cancel against the factor in the first, so the optimum generally depends on the law of $N$. The two endpoints are count-free for different reasons. At $\lambda=0$ the reserve $v_0$ is optimal whenever there is positive probability of at least two bidders, whereas $\lambda=1$ recovers Proposition \ref{prop:policyid}(iii).

The bid log does not record the negotiated outcomes of unmet auctions, so $\lambda$ cannot be evaluated in these data. Wholesale used-vehicle auctions employ the same combination of a secret reserve and post-auction bargaining and do record the negotiation, and sellers in that setting capture a substantial but incomplete share of the available gains \citep{larsen2021efficiency}. If that result carries over, $\lambda$ lies in the interior of its range here, and the reserves of Section \ref{sec:results} are an upper bound on the optimum under the platform's rule. The broader secret-reserve literature considers a further channel that Assumption \ref{ass:mechanism}(a) sets aside, in which the reserve moves bidding through beliefs or entry \citep{elyakime1994first,vincent1995bidding,eklof2003open,foster2022semi}. The public ``No Reserve'' label is the one piece of reserve information that bidders observe, and it enters the fitted model only through the listing text.

\subsection{Full-Bargaining Benchmark for the Policy Disagreements}\label{sec:bargaining-benchmark}

Table \ref{tab:bargaining-benchmark} reports the empirical continuation benchmark under Assumption \ref{ass:mechanism}(b) for the \ResPolNLow{} reserve listings whose final bids fell below $r^\ast$ even though the seller's reserve was met. Consistent with the continuous-bid approximation above, the calculation sets $V_{(2)}=b_{(1)}$. Raising the reserve then leaves the bid path unchanged and sends the seller and the high bidder to negotiation, where the seller offers $r^\ast$ and the bidder accepts with conditional probability $S(r^\ast\mid z)/S(b_{(1)}\mid z)$. The expected payoff difference is
\[
v_0+(r^\ast-v_0)\frac{S(r^\ast\mid z)}{S(b_{(1)}\mid z)}-b_{(1)}.
\]
Its median is \$\ResPolMedianGain{} and its total is \$\ResPolTotalGain{} million, or \ResPolGainShare\% of realized revenue in this cell. Across the multipliers $c=0.40$ and $c=0.50$, which bracket the calibrated value $c^\ast=\ResCStar$, the total ranges from \$\ResPolTotalCLow{} million to \$\ResPolTotalCHigh{} million. Because the bid log does not record the negotiation outcome, this is a full-bargaining benchmark and not an observed gain from changing the incumbent policy.

\begin{table}[!htbp]
  \centering
  \caption{Full-Bargaining Benchmark for Auctions Clearing Below the Model's Reserve}
  \label{tab:bargaining-benchmark}
  \small
\begin{tabular}{lr}
\toprule
Auctions & 4,194 \\
Median gain (\$) & 1,239 \\
Mean gain (\$) & 2,840 \\
Median gain, share of the winning bid (\%) & 6.2 \\
Median acceptance probability of the offer at $r^\ast$ (\%) & 48.1 \\
Total gain (\$ million) & 11.9 \\
Realized revenue on these auctions (\$ million) & 101.0 \\
Total gain, share of realized revenue (\%) & 11.8 \\
\bottomrule
\end{tabular}

  \par\smallskip
  \begin{minipage}{0.92\textwidth}
    \footnotesize \textit{Note.} The sample contains reserve listings whose final bid met the seller's reserve but fell below $r^\ast$. No-reserve listings are excluded. The calculation holds the bid path fixed and uses the continuous-bid approximation $V_{(2)}=b_{(1)}$ despite observed bid increments. Under Assumption \ref{ass:mechanism}(b), the seller offers $r^\ast$ after the auction and the high bidder accepts with conditional probability $S(r^\ast\mid z)/S(b_{(1)}\mid z)$. Every auction is evaluated at $v_0=c^\ast\widehat{\E}[B_{(2)}\mid\hat z]$.
  \end{minipage}
\end{table}

\section{The Value of Fine-Tuning and Specification Checks}\label{app:robustness}


\subsection{The Value of Structural Fine-Tuning}\label{sec:head-to-head}

Fine-tuning the embedding model on the structural criterion adds a stage that the pre-trained index of the same embedding model does not require. We compare the two indices on the held-out $T_{\mathrm{II}}$ auctions along two criteria, the error in a dollar prediction of the second-highest bid and the likelihood on which the estimator rests (Table \ref{tab:fit}, Panel A).

Two held-out auctions fall outside the fine-tuned model's GPD domain, while none falls outside the pre-trained model's domain. We report these failures separately and compare both models on the remaining \ResComparisonValidN{} auctions. On this common support, fine-tuning raises mean log likelihood by \ResHeldOutDiff{} nats, with a paired $t$ statistic of \ResHeldOutT. However, this conditional comparison does not establish improved likelihood performance over the full held-out sample.

On the same common support, the held-out median absolute error of the predicted second-highest bid falls from \$\ResDollarMAEPT{} under the pre-trained index to \$\ResDollarMAEFT{} under the fine-tuned index. The paired mean reduction is \$\ResDollarPairedMean{} with a standard error of \$\ResDollarPairedSE, and the fine-tuned index is closer on \ResDollarShareCloser\% of auctions. The fine-tuned index also more than doubles the out-of-sample log $R^2$ of the seller-value anchor, from \ResAnchorRsqPre{} to \ResAnchorRsq, evaluated on all \ResNDB{} held-out auctions.

Three features qualify these gains. The comparison fits the same projection-head family under the same multistart protocol on $T_{\mathrm{I}}$ under each index, scores both fits on $T_{\mathrm{II}}$, and treats auctions as independent in the paired standard errors, so the experimental difference is the index. The fine-tuned index uses $T_{\mathrm{I}}$ outcomes by construction, and its in-sample to held-out gap of \ResGapFT{} nats, against \ResGapPT{} for the pre-trained index, reflects that use as well as ordinary sample variation. Both indices over-predict the second-highest bid at the median, by \ResDollarOverFT\% and \ResDollarOverPT\%, so the dollar error ranks the indices within one tail specification and does not validate either. Finally, sellers and vehicle types recur across halves because the split is at the auction level, so the measured advantage may be optimistic relative to a split by vehicle identity.

The rank diagnostics use $U_{jt}=\hat S(B_{(j),t}\mid z_t)/\hat S(B_{(k_t),t}\mid z_t)$ and the Rosenblatt components
\begin{equation}
 W_{jt}=\left(\frac{U_{jt}}{U_{j+1,t}}\right)^j,\quad j=2,\ldots,k_t-2,
 \qquad W_{k_t-1,t}=U_{k_t-1,t}^{k_t-1}.
 \label{eq:rosenblatt}
\end{equation}

\begin{table}[!htbp]
  \centering
  \caption{Representation Quality and Specification Fit}
  \label{tab:fit}
  \small
\begin{tabular*}{\textwidth}{@{\extracolsep{\fill}}lrr@{}}
\toprule
& Fine-tuned index & Pre-trained index \\
\midrule
\multicolumn{3}{@{}l}{\emph{Panel A. Held-out fit on the inference half}} \\
Held-out auctions outside the GPD domain & 2 & 0 \\
Held-out mean log likelihood (nats) & 2.5596 & 2.5102 \\
In-sample mean log likelihood (nats) & 2.6484 & 2.5305 \\
In-sample to held-out gap (nats) & 0.089 & 0.020 \\
Seller-value anchor, out-of-sample log $R^2$ & 0.634 & 0.296 \\
Difference in held-out log likelihood (nats) & \multicolumn{2}{r}{0.0495} \\
Paired standard error (nats) & \multicolumn{2}{r}{0.0029} \\
Paired $t$ statistic & \multicolumn{2}{r}{17.2} \\
Share of auctions favoring the fine-tuned index (\%) & \multicolumn{2}{r}{61.9} \\
Median absolute error, predicted second-highest bid (\$) & 4,522 & 5,080 \\
Median absolute relative error (\%) & 17.9 & 20.7 \\
Median ratio of predicted to realized second-highest bid & 1.170 & 1.200 \\
Paired reduction in absolute error (\$) & \multicolumn{2}{r}{446} \\
Paired standard error (\$) & \multicolumn{2}{r}{31} \\
Share of auctions where the fine-tuned index is closer (\%) & \multicolumn{2}{r}{59.6} \\
\midrule
\end{tabular*}

\begin{tabular*}{\textwidth}{@{\extracolsep{\fill}}lrrr@{}}
\multicolumn{4}{@{}l}{\emph{Panel B. Rank-ordered probability integral transform, fine-tuned index}} \\
\midrule
Component & Contributions & Mean & Kolmogorov--Smirnov \\
\midrule
$j=2$ & 32,853 & 0.617 & 0.188 \\
$j=3$ & 27,670 & 0.514 & 0.061 \\
$j=4$ & 21,812 & 0.474 & 0.080 \\
$j=5$ & 16,091 & 0.452 & 0.100 \\
$j=6$ & 11,199 & 0.432 & 0.116 \\
$j=7$ & 7,394 & 0.415 & 0.136 \\
$j=8$ & 4,666 & 0.404 & 0.151 \\
All components pooled & 127,715 & 0.503 & 0.053 \\
Last component only & 32,853 & 0.462 & 0.074 \\
\midrule
\end{tabular*}

\begin{tabular*}{\textwidth}{@{\extracolsep{\fill}}lrrrr@{}}
\multicolumn{5}{@{}l}{\emph{Panel C. Reserve movement under alternative Stage II fits at the calibrated seller value}} \\
\midrule
Refit & Auctions & Movement (\%) & Shift (\%) & Rank correlation \\
\midrule
Retention window, one hour & 29,883 & 3.6 & $-$3.8 & 0.99990 \\
Retention window, 24 hours & 32,853 & 5.2 & 5.6 & 0.99986 \\
\multicolumn{5}{@{}l}{Two-piece inverse hazard, knot at} \\
\quad the 25th percentile threshold (\$9,850) & 32,853 & 0.4 & 0.2 & 0.99990 \\
\quad the 50th percentile threshold (\$19,000) & 32,853 & 0.6 & 0.4 & 0.99980 \\
\quad the 75th percentile threshold (\$38,500) & 32,853 & 0.5 & 0.1 & 0.99992 \\
\quad the price anchor & 32,853 & 0.4 & $-$0.0 & 0.99995 \\
\bottomrule
\end{tabular*}

  \par\smallskip
  \begin{minipage}{0.92\textwidth}
    \footnotesize \textit{Note.} Panel A fits the same projection-head family on $T_{\mathrm{I}}$ under each index. Domain failures are counted across all \ResNDB{} auctions of $T_{\mathrm{II}}$. Held-out likelihood and dollar-prediction statistics use the \ResComparisonValidN{} auctions on which both models have valid support. The anchor's out-of-sample log $R^2$ uses all \ResNDB{} auctions. Paired standard errors treat auctions as independent. The predicted second-highest bid is its conditional median given the retained threshold and depth. Panel B reports the Rosenblatt components of equation \eqref{eq:rosenblatt} under the fine-tuned index by rank $j$. With fixed true parameters, Lemma \ref{lem:song} and the fitted tail imply independent uniforms. Panel C refits Stage II on the bids retained under alternative windows and under a two-piece inverse hazard whose slope may change above the stated knot, holding the Stage I representation fixed. Movement is the median absolute relative change in individual reserves, shift is the relative change in the median reserve, and the rank correlation is with the reserves under the baseline specification.
  \end{minipage}
\end{table}

\section{Computational Details}\label{app:computation}

This appendix records the optimization, solver, and hardware settings behind Section \ref{sec:results}. The table collects the selected production settings. The accompanying code records the random seeds, software environment, optimizer tolerances, and structured-start protocol used to reproduce the estimates.

\subsection{Stage I Selection}

Holding the index dimension at 64, selection on the within-$T_{\mathrm{I}}$ validation slice favors a rank-16 adapter at an embedding model learning rate of $10^{-4}$, which reaches a validation negative log likelihood of $\ResStageOneBest$ at epoch \ResStageOneBestEpoch. Halving the embedding model's learning rate or doubling the adapter rank raises the validation negative log likelihood by \ResStageOneHalfLRLoss{} and \ResStageOneRankLoss{} nats. No completed run has a scale parameter at its floor or a non-positive likelihood argument.

\subsection{Stage II Convergence and Standard Errors}

The Stage II criterion attains $\ResStageTwoObjective$ on the 32,886 inference-half auctions. Of \ResStageTwoStarts{} starts, \ResStageTwoConverged{} converge, the relative spread of the criterion across converged starts is \ResStageTwoSpread, and the converged solutions agree in function space, with a maximum relative reserve movement of \ResStageTwoReserveMovement{} over a \ResStageTwoAuditRows-auction audit set. The best start is the structured start initialized at the Stage I head $\hat{\gamma}_{\mathrm{I}}$. Domain violations are zero on $T_{\mathrm{II}}$ and on the \ResNDA{} representation-half auctions, which Stage II never used.

The descriptive comparison of Appendix \ref{sec:head-to-head} instead fits a head on $T_{\mathrm{I}}$ under each index and scores it on $T_{\mathrm{II}}$. Both comparison fits use the 32 Sobol starts and the two structured starts of the protocol and omit the third structured start, the Stage I head, which has no counterpart for the pre-trained index. The pre-trained fit \ResDollarPTGateStatus{} with \ResDollarPTConverged{} of \ResDollarPTStarts{} starts converged. The fine-tuned fit converges on \ResDollarFTConverged{} of \ResDollarFTStarts{} starts, and its converged starts agree, but it \ResDollarFTGateStatus. Among the \ResDollarFTHoldoutRows{} held-out auctions, one has a negative fitted scale and a different auction has an observed bid above the fitted upper endpoint, giving \ResDollarFTHoldoutViolations{} auctions outside the GPD domain. Their clamped numerical scores are not literal GPD log densities and are excluded from both models' likelihood and dollar-prediction comparisons. We report the two failures separately and calculate paired comparisons on the remaining \ResComparisonValidN{} auctions. These exclusions affect only the descriptive comparison. None of these comparison fits enters the reported reserve estimates.

Standard errors follow the sandwich construction of Section \ref{sec:training} on the stacked 258-coordinate system that carries the 130 projection-head parameters with the 128 standardization coordinates, so uncertainty in the standardization propagates. The per-record score averages to the stored mean equation with a maximum gap of \ResScoreGap, the Jacobian is full rank with condition number \ResJacobianCond, and the analytic gradient of the reserve used by the delta method agrees with automatic differentiation to a relative gap of \ResGradientGap.

\subsection{Support Band and Ancillary Retention Sensitivity}

The support flag of Section \ref{sec:reserve-estimates} compares each reserve with the band between the 5th and 95th percentiles of the retained thresholds of its 512 nearest inference-half listings in the index, whose lower endpoint estimates $\underline{t}(d)$ of Section \ref{sec:song} at $\alpha=0.05$. Holding $c^\ast$ fixed and preserving incumbent outcomes for the \ResSupportBelow{} listings below the estimated identification cutoff lowers the net automatic-sale revenue difference by \$42,450. The median lower endpoint is \$\ResBandMedianLower{} against a median own retained threshold of \$\ResOwnThresholdMedian, and \ResReserveBelowOwnThresholdShare\% of reserves lie below the auction's own threshold. Narrowing the band to the interquartile range at the same 512 neighbors moves the median endpoints to \$\ResBandNarrowLower{} and \$\ResBandNarrowUpper{} and the in-band share from \ResBandWideInShare\% to \ResBandNarrowInShare\%, with \ResBandNarrowBelow{} reserves below the band and \ResBandNarrowAbove{} above it. The neighbor count has a smaller effect. At the wider band the in-band share is \ResBandNoneTwoEightInShare\% with 128 neighbors and \ResBandNtwoZeroFourEightInShare\% with 2,048.

The first Rosenblatt component $W_{2t}$ drifts with the observed count of active bidders as it does with the retained depth. Its mean rises from \ResWtwoCountLow{} in auctions with three to eight observed bidders to \ResWtwoCountHigh{} in auctions with nineteen or more, a Spearman correlation of \ResWtwoSpearmanCount, against \ResWtwoSpearmanK{} with depth. Under the independence of valuations and market size maintained in Section \ref{sec:distribution}, the dependence on the count should also be absent. Depth and observed count are themselves correlated at \ResCountDepthCorr, so the two drifts cannot be separated in these data, and both are joint tests of the restriction and the tail specification.

\subsection{Settings}

Table \ref{tab:computational-settings} summarizes the settings of both stages and the compute they consumed.

\begin{table}[htbp]
  \centering
  \caption{Computational Settings}
  \label{tab:computational-settings}
  \small
  \begin{tabular}{@{}lcp{0.50\textwidth}@{}}
    \toprule
    & Pre-trained benchmark & Fine-tuned index \\
    \midrule
    \multicolumn{3}{@{}l}{\emph{Stage I (representation)}} \\
    Sample & --- & $T_{\mathrm{I}}$ alone \\
    Objective & --- & Structural criterion of Section \ref{sec:training}, no auxiliary predictive targets \\
    Solver & --- & AdamW, at most 8 epochs, early stopping \\
    Learning rates & --- & Adapter $10^{-4}$, projection head $5\times10^{-3}$, 3\% warmup, cosine decay to $0.1$ of peak \\
    Batching & --- & Effective batch 256 auctions, gradient clipping 1.0, 32,768-token micro-batches, gradient checkpointing \\
    Adapter & --- & LoRA rank 16, scaling 32, dropout 0.05, on $q$, $k$, $v$, and $o$ projections \\
    Adapter capacity & --- & 4,587,520 trainable parameters, 0.77\% of the embedding model \\
    Selection & --- & Validation NLL within $T_{\mathrm{I}}$ (10\% slice, held fixed) over adapter rank, model learning rate, epoch, and initialization \\
    \addlinespace
    \multicolumn{3}{@{}l}{\emph{Stage II (inference)}} \\
    Sample & \multicolumn{2}{l@{}}{$T_{\mathrm{II}}$ alone, index cached as input, no gradient to $\mathcal{M}$} \\
    Solver & \multicolumn{2}{l@{}}{L-BFGS-B, full batch, float64, 35 starts (32 Sobol, 3 structured)} \\
    Production gate & \multicolumn{2}{l@{}}{Converged starts must agree in function space over the fitted region} \\
    \addlinespace
    \multicolumn{3}{@{}l}{\emph{Hardware and compute}} \\
    Representation & \multicolumn{2}{l@{}}{One bfloat16 pass, NVIDIA H100 NVL} \\
    Stage I training & --- & H100, 64 minutes to the selected epoch \\
    Structural estimation & \multicolumn{2}{l@{}}{CPU float64, about 6 minutes per fit} \\
    \bottomrule
  \end{tabular}
  \par\smallskip
  \begin{minipage}{0.92\textwidth}
    \footnotesize \textit{Note:} Stage I produces the fine-tuned index. Stage II applies to the fine-tuned index and to the pre-trained benchmark, which differ only in the index supplied to the projection head. Compute figures are for the reported production runs.
  \end{minipage}
\end{table}

\clearpage
\bibliographystyle{plainnat}
\bibliography{refs}
\end{document}